\documentclass[10pt,twocolumn,letterpaper]{article}

\usepackage[pagenumbers]{cvpr}

\usepackage[dvipsnames]{xcolor}

\usepackage{comment}

\newcommand{\shortname}{\textit{SynShot}\xspace}
\newcommand{\longname}{Synthetic Prior for Few-Shot Drivable Head Avatar Inversion\xspace}

\definecolor{cvprblue}{rgb}{0.21,0.49,0.74}
\usepackage[pagebackref,breaklinks,colorlinks,citecolor=cvprblue]{hyperref}
\usepackage[normalem]{ulem}
\usepackage{soul}
\usepackage{balance}

\renewcommand{\paragraph}[1]{\noindent\textbf{#1}}

\title{\longname}

\author{
Wojciech Zielonka\textsuperscript{1, 2, 3*} \quad Stephan J. Garbin\textsuperscript{3} \quad Alexandros Lattas\textsuperscript{3} \\ \quad George Kopanas\textsuperscript{3} \quad Paulo Gotardo\textsuperscript{3} \quad Thabo Beeler\textsuperscript{3} \quad Justus Thies\textsuperscript{1, 2} \quad Timo Bolkart\textsuperscript{3} \\ \\
\textsuperscript{1}Max Planck Institute for Intelligent Systems, Tübingen, Germany \\
\textsuperscript{2}Technical University of Darmstadt \quad
\textsuperscript{3}Google \\
\url{https://zielon.github.io/synshot/}
}

\begin{document}

\twocolumn[{%
\renewcommand\twocolumn[1][]{#1}%
\maketitle
\begin{center}
    \centering
    \captionsetup{type=figure}
    \includegraphics[width=0.95\textwidth]{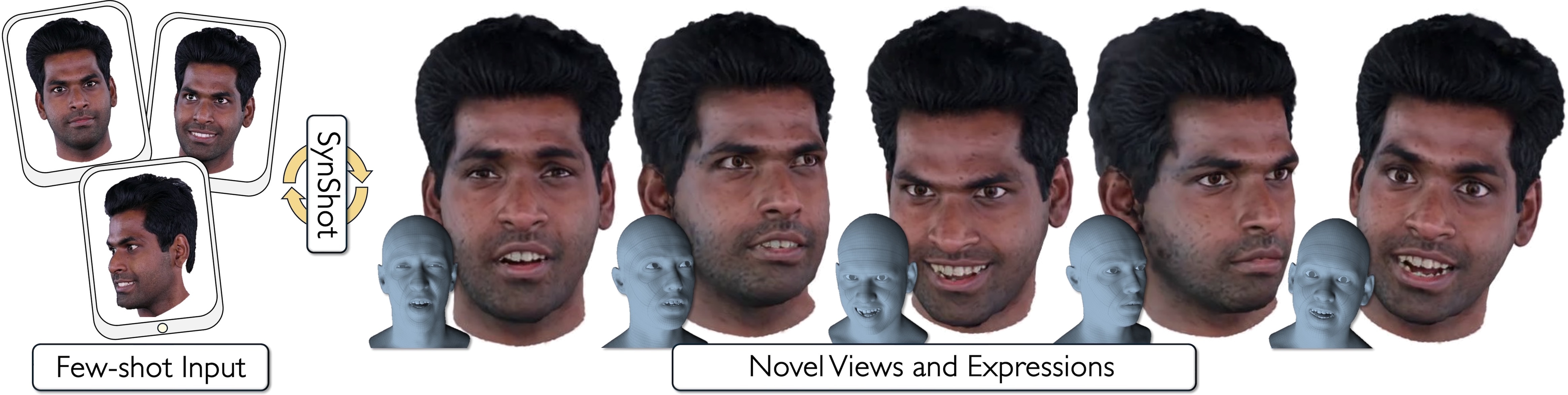}
    \caption{
    Given a few input images (left), \textbf{\shortname} generates a personalized 3D Gaussian avatar that renders from new viewpoints and unseen expressions (right).
    To compensate for the missing information in the input images, we leverage a generative Gaussian head avatar trained on a diverse synthetic head dataset as a 3D prior.
    %
    %
    }
    \label{fig:teaser}
\end{center}
}]

\let\thefootnote\relax\footnotetext{$^*$Work done while WZ was interning at Google in Zurich, Switzerland}\par

\newcommand{\texturemap}{\pmb{x}_{\mathrm{tex}}\xspace}
\newcommand{\texturemapdim}{\mathbb{R}^{H \times W \times 3}\xspace}

\newcommand{\vertexmap}{\pmb{x}_{\mathrm{verts}}\xspace}
\newcommand{\vertexmapdim}{\mathbb{R}^{H \times W \times 3}\xspace}

\newcommand{\expressionmap}{\pmb{x}_{\mathrm{exp}}\xspace}
\newcommand{\expressionmapdim}{\mathbb{R}^{H \times W \times 3}\xspace}

\newcommand{\featuredim}{F\xspace}
\newcommand{\outfeaturemap}{\hat{\pmb{x}}_{\mathrm{feat}}\xspace}
\newcommand{\outfeaturemapdim}{\mathbb{R}^{H \times W \times \featuredim}\xspace}
\newcommand{\outvertexmap}{\hat{\pmb{x}}_{\mathrm{verts}}\xspace}
\newcommand{\outtexturemap}{\hat{\pmb{x}}_{\mathrm{tex}}\xspace}
\newcommand{\outexpressionmap}{\hat{\pmb{x}}_{\mathrm{expr}}\xspace}
\newcommand{\sampledfeatures}{\pmb{s}\xspace}

\newcommand{\identityencoder}{\pmb{E}_{\mathrm{id}}\xspace}
\newcommand{\expressionencoder}{\pmb{E}_{\mathrm{expr}}\xspace}

\newcommand{\quantization}{\mathbf{q}\xspace}

\newcommand{\featuredecoder}{\pmb{D}_{\mathrm{feat}}\xspace}
\newcommand{\identitydecoder}{\pmb{D}_{\mathrm{id}}\xspace}
\newcommand{\expressiondecoder}{\pmb{D}_{\mathrm{expr}}\xspace}

\newcommand{\latentcode}{\pmb{z}\xspace}
\newcommand{\identitycode}{\latentcode_{\mathrm{id}}\xspace}
\newcommand{\identitycodedim}{\mathbb{R}^{h \times w \times n_{\mathrm{id}}}\xspace}
\newcommand{\expressioncode}{\latentcode_{\mathrm{expr}}\xspace}
\newcommand{\expressioncodedim}{\mathbb{R}^{h \times w \times n_{\mathrm{expr}}}\xspace}

\newcommand{\colorregressor}{\pmb{R}_{color}\xspace}
\newcommand{\gaussianregressor}{\pmb{R}_{gauss}\xspace}

\newcommand{\position}{\pmb{\phi}\xspace}
\newcommand{\rotation}{\pmb{\theta}\xspace}
\newcommand{\scale}{\pmb{\sigma}\xspace}
\newcommand{\opacity}{\pmb{\alpha}\xspace}
\newcommand{\sphericalharmonics}{\pmb{h}\xspace}

\newcommand{\sampling}{\mathcal{B}\xspace}
\newcommand{\splat}{\bar{\mathcal{I}}\xspace}
\newcommand{\image}{\mathcal{I}\xspace}

\begin{abstract}

We present \textbf{\shortname}, a novel method for the few-shot inversion of a drivable head avatar based on a synthetic prior.
We tackle three major challenges.
First, training a controllable 3D generative network requires a large number of diverse sequences, for which pairs of images and high-quality tracked meshes are not always available.
Second, the use of real data is strictly regulated (e.g., under the \textbf{G}eneral \textbf{D}ata \textbf{P}rotection \textbf{R}egulation, which mandates frequent deletion of models and data to accommodate a situation when participant's consent is withdrawn). Synthetic data, free from these constraints, is an appealing alternative.
Third, state-of-the-art monocular avatar models struggle to generalize to new views and expressions, lacking a strong prior and often overfitting to a specific viewpoint distribution.
Inspired by machine learning models trained solely on synthetic data, we propose a method that learns a prior model from a large dataset of synthetic heads with diverse identities, expressions, and viewpoints.
With few input images, \shortname fine-tunes the pretrained synthetic prior to bridge the domain gap, modeling a photorealistic head avatar that generalizes to novel expressions and viewpoints.
We model the head avatar using 3D Gaussian splatting and a convolutional encoder-decoder that outputs Gaussian parameters in UV texture space. 
To account for the different modeling complexities over parts of the head (e.g., skin {\em vs} hair), we embed the prior with explicit control for upsampling the number of per-part primitives. 
Compared to SOTA monocular and GAN-based methods, \shortname significantly improves novel view and expression synthesis.

\end{abstract}    
\section{Introduction}
\label{sec:intro}

The ability to build high-fidelity drivable digital avatars is a key enabler for virtual reality (VR) and mixed reality (MR) applications.
However, creating photorealistic human head models \cite{Oleg2009emily,seymour2017mike} using traditional rendering assets requires sophisticated data capture and significant manual cleanup, which is time-consuming and expensive.

The recent advancements in learning-based methods and radiance fields~\cite{mildenhall2020nerf,kerbl2023gaussian} have simplified the avatar creation process, leading to impressive progress in quality and democratization of neural head avatars
\cite{Ma2021,wang2023styleavatar,Gafni2021nerface}.
Such progress is particularly noticeable in enhancing control through lightweight animation \cite{Qian2024gaussianavatars,zielonka2024gem,xiang2024flashavatar}, and reducing training time to a few minutes \cite{Zielonka2022InstantVH}.
These methods are trained on multi-view \cite{Ma2021,wang2023styleavatar,Qian2024gaussianavatars} or single-view videos \cite{Gafni2021nerface,xiang2024flashavatar,Chen2024MonoGaussianAvatar,Zielonka2022InstantVH}, typically requiring hundreds to thousands of video frames.
Processing such datasets is complex and error-prone as most methods require tracking a coarse head mesh across all frames, which is typically done by fitting a 3D morphable model \cite{bfm09,Li2017flame} to the image. 
A further limitation of existing personalized head avatars is their poor generalization to facial expressions and camera viewpoints not captured in the set of input images.

Another recent body of work addresses the problem of building 3D head avatars from one or few input images, \cite{Yu2024One2Avatar,Chen2024,chu2024generalizableanimatablegaussianhead}. However, their rendering quality and fidelity are typically lower than those of methods trained on large datasets (e.g., \cite{Qian2024gaussianavatars,zielonka2024gem}). To improve quality, some methods \cite{zheng2024headgap,xu2024gphmv2, xu2023gphm} first learn a multi-identity head model that is used as prior when optimizing for the personalized avatar. Training these head priors requires a large-scale multi-view image dataset that is expensive and time-consuming to capture. Moreover, managing real data under protection laws like \textbf{GDPR} is cumbersome for experimentation and maintenance, as users must periodically (e.g., every 30 days) delete all dataset derivatives and trained models, allowing dataset participants to be removed from both if needed.
Alternatively, the FFHQ dataset \cite{Karras_2019_CVPR} may be employed, with 4D GAN-based methods \cite{invertavatar, sun2023next3d, deng2024portrait4d} constructing an inversion prior from it. However, these approaches tend to exhibit artifacts during novel view synthesis and struggle with preserving identity.
In summary, the expressive power of this prior is strongly influenced by: the training data diversity (\eg, ethnicity, age, facial features, expressions), the multi-view capture hardware setup (\ie, lighting, view-density, calibration quality, frame-rate), and the quality of the data pre-processing (\eg, mesh tracking, background masking).

In contrast to the previous work that focuses on expensive and cumbersome real data, we overcome these limitations and propose \shortname, a new method that builds a prior solely on synthetic data and adapts to a real test subject requiring only a few input images.
Building on the success of ML models trained on synthetic data for tasks like 3D face regression \cite{Sela2017}, 2D landmark prediction \cite{wood2021fake}, rigid face alignment \cite{Bednarik2024_EG}, and few-shot head reconstruction \cite{wang2023rodin, zhang2024rodinhd, buehler2024cafca}, \shortname is trained solely on a large synthetic dataset generated from 3DMM samples and diverse assets.
Synthetic data offers complete control over dataset creation to meet size and diversity needs for training an expressive head prior, eliminating the need for costly capture hardware and addressing privacy concerns with real subjects.
The benefits brought by synthetic data come at the cost of having to handle the domain gap between the trained head prior and real images captured ``in the wild''. To effectively bridge this gap, we first fit the synthetic prior to real images and then fine-tune the prior weights to the real data using the pivotal tuning strategy proposed in \cite{pivotaltuning}.
With as few as three input images, \shortname reconstructs a photorealistic head avatar that generalizes to novel expressions and camera viewpoints (Fig.~\ref{fig:teaser}). The results show that our method outperforms state-of-the-art personalized monocular methods \cite{Zielonka2022InstantVH, xiang2024flashavatar, shao2024splattingavatar} trained on thousands of images each. Our method represents head avatars using 3D Gaussian primitives~\cite{kerbl2023gaussian}, where Gaussian parameters are generated by a convolutional architecture in UV space \cite{li2024animatablegaussians, zielonka2024gem, saito2024rgca}. 

\smallskip
\noindent
In summary, our key contributions are:
\begin{enumerate}
\item A generative method based on a convolutional encoder-decoder architecture that is trained on extensive synthetic data only to produce controllable 3D head avatars.
\item A reconstruction scheme that adapts and fine-tunes a pretrained generative model on a few real images to create a personalized, photorealistic 3D head avatar.
\end{enumerate}

\section{Related Work}
\label{sec:related work}
\begin{figure*}[ht!]
  \centering
  \includegraphics[width=1.0\linewidth]{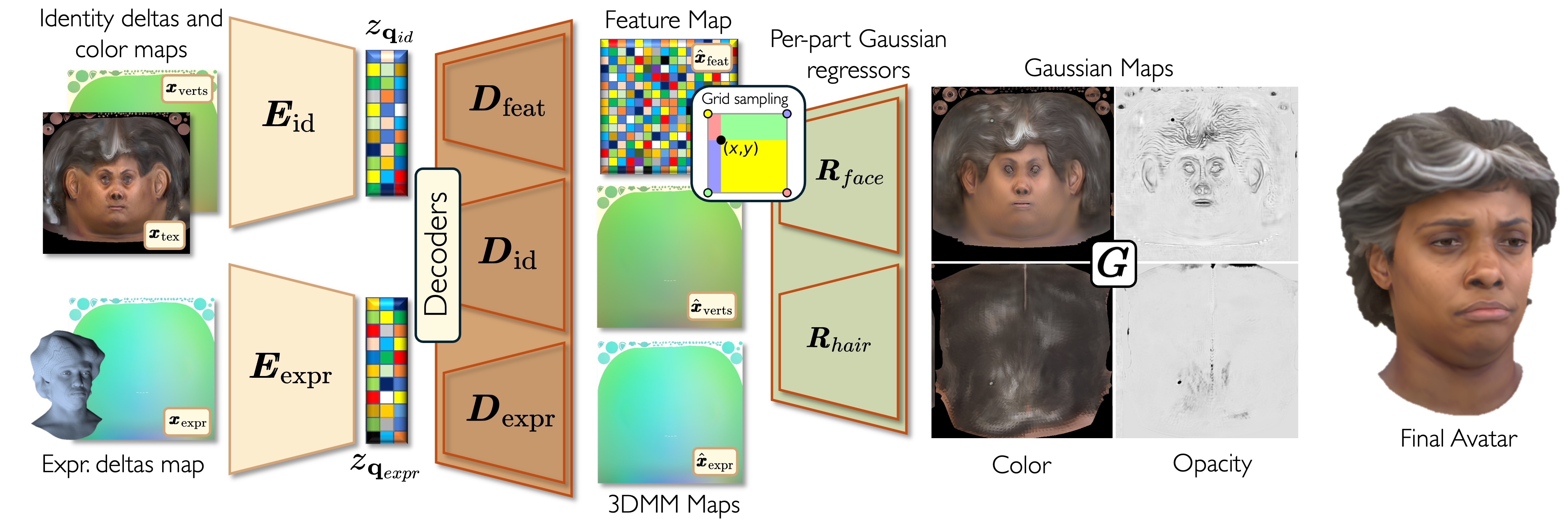}
  \caption{\textbf{Pipeline overview.} Given an extracted texture $\texturemap$, rasterized position map $\vertexmap$, and PCA expression deltas $\expressionmap$ our network utilized VQ-VAE to jointly optimize for two latent space $\expressioncode$ and $\identitycode$. The VQ-VAE decoders predict feature map $\featuredecoder(\quantization(\identitycode), \quantization(\expressioncode)) \rightarrow \outfeaturemap$, identity and color maps $\identitydecoder(\quantization(\identitycode))\rightarrow\{\outtexturemap, \outvertexmap\}$, and expression deltas $\expressiondecoder(\quantization(\expressioncode)) \rightarrow \outexpressionmap$. Finally, bilinearly sampled maps are passed to per-part regressors $\colorregressor$ and $\gaussianregressor$ to obtain primitives to rasterize.}
  \label{fig:overview}
\end{figure*}
%
\paragraph{Few-shot Head Avatars.}
%
%
3D Morphable Models (3DMM) \cite{BlanzVetter1999, bfm09, Li2017flame, Egger20203DMM} have long been used for creating facial avatars. When paired with generative models for textures \cite{gecer2019ganfit, lattas2020avatarme, lattas2021avatarme++, luo2021normalized}, 3DMMs can be optimized from in-the-wild images. Techniques such as inverse rendering \cite{dib2024mosar}, diffusion-based inpainting \cite{papantoniou2023relightify}, and pivotal-tuning \cite{pivotaltuning, lattas2023fitme} are used to disentangle appearance from identity.
%
Neural radiance fields (NeRF) \cite{mildenhall2020nerf} and 3D Gaussian representations (3DGS) \cite{kerbl2023gaussian} have also been widely used for avatar reconstruction. EG3D \cite{chan2022efficient} employs features on tri-planes, enabling consistent 3D face generation and inversion from in-the-wild images. PanoHead \cite{panohead2023} extends EG3D through tri-grids to achieve a full 360-degree generation of static human heads. Gaussian3Diff \cite{lan2023gaussian3diff} further improves quality by replacing neural features with 3D Gaussians.
%
Rodin \cite{wang2023rodin} and RodinHD \cite{zhang2024rodinhd} leverage an extensive dataset of synthetic humans to train a triplane-based avatar generator used to invert in-the-wild images; however, the results remain confined to the synthetic domain and avatars are not drivable.

Diner \cite{prinzler2022diner} incorporates depth information, while Preface \cite{buehler2023preface} trains a volumetric prior on synthetic human data and fits it to a few input images to match a subject's likeness. Cafca \cite{buehler2024cafca} extends Preface to better generalize to static but arbitrary facial expressions. In contrast, our method not only bridges the domain gap from synthetic to real but also produces animatable avatars. MofaNeRF \cite{zhuang2022mofanerf} and NeRFace \cite{Gafni2021nerface} condition NeRFs on expression (and shape) codes, while HeadNeRF \cite{hong2021headnerf} similarly embeds NeRFs into parametric models. Portrait4D \cite{deng2024portrait4d} introduces one-shot 4D head synthesis using a transformer-based animatable triplane reconstructor built on the EG3D \cite{chan2022efficient}. Next3D \cite{sun2023next3d} employs GAN-based neural textures embedded on a parametric mesh; however, it suffers from inversion problems. InvertAvatar \cite{invertavatar} tackles the shortcomings of Next3D and further refines avatar inversion using few-shot images. Despite relatively good frontal performance, these GAN-based methods often exhibit artifacts, such as identity changes, in novel view synthesis.

Recent methods by Xu \etal \cite{xu2023gphm} are conceptually similar to HeadNeRF and MofaNeRF; however, instead of embedding NeRF \cite{mildenhall2020nerf} on a mesh, they employ 3DGS \cite{kerbl2023gaussian}. GPHM \cite{xu2023gphm} uses a series of MLPs to generate Gaussian primitives attached to a parametric model, enabling expression control and inversion, though it conditions only the avatar's shape. 
GPHMv2 \cite{xu2024gphmv2} extends GPHM with a dynamic module for improved reenactment control and a larger dataset, further enhancing quality. HeadGAP \cite{zheng2024headgap} also models avatars using MLPs, utilizing part-based features and additional color conditioning to improve quality. While these methods embed primitives directly on the mesh surface, our approach explicitly learns the primitive parameters by modeling their distribution via a VQ-VAE \cite{van2017neural}, eliminating the need for a guiding mesh during the test time as the shape is captured within our latent space.

\paragraph{Multi-view Personalized Avatars.}
Volumetric primitives, combined with multi-view training, are highly effective for modeling human heads \cite{zielonka2024gem, saito2024rgca, xu2023gaussianheadavatar, giebenhain2024npga, kirschstein2023diffusionavatars, kabadayi24ganavatar, teotia24hq3d, teotia24gaussianheads} as they capture intricate details like hair and subsurface scattering \cite{sarkar2023litnerf}.
VolTeMorph \cite{VolTeMorph} embeds a NeRF within tetrahedral cages that guide volumetric deformation. Qian \etal \cite{Qian2024gaussianavatars} attach Gaussian primitives to 3DMM triangles, whose local rotations and stretch deform the Gaussians without requiring neural networks. Xu \etal \cite{xu2023gaussianheadavatar} and Giebenhain \etal \cite{giebenhain2024npga} extend that work to further predict corrective fields over the Gaussians; rather than colors, they splat features that are translated into color by an image-space CNN \cite{thies2019deferred}. Lombardi \etal \cite{Lombardi21mvp} position 3D voxels with RGB and opacity values at the vertices of a head mesh, using ray tracing for volumetric integration. Saito \etal \cite{saito2024rgca} improve quality by replacing voxel primitives with 3D Gaussians and applying rasterization. Our VQ-GAN training aligns with these principles {\em for few-shot capture}, as we supervise the process using a hybrid mesh-primitive approach to model the generative distribution.

\paragraph{Monocular Personalized Avatars.}
Monocular methods often rely on a strong 3DMM prior, as recovering a 3D shape from a 2D image is an inherently under-constrained problem. Face2Face
\cite{thies2016face} was a seminal work that enabled real-time reconstruction and animation of a parametric model. However, it lacks detailed hair representation and relies on low-frequency PCA texture models, which significantly affects quality. This limitation has led to the rise of neural avatars based on NeRF \cite{wang2023styleavatar, Yu2024One2Avatar, xu2023latentavatar, zheng2022imavatar, xu2023avatarmav, chu2024gagavatar, zheng2022imavatar, zheng2023pointavatar, feng2023delta, gao2022reconstructing, Gafni2021nerface} and later on 3D Gaussian primitives \cite{xiang2024flashavatar, shao2024splattingavatar, kirschstein2024gghead, Chen2024MonoGaussianAvatar}. INSTA \cite{Zielonka2022InstantVH} applies triangle deformation gradients \cite{Sumner2004DeformationTF} to each NeRF sample based on proximity to the nearest triangle, enabling avatar animation. This approach has been adapted to 3D Gaussian Splatting (3DGS) by methods like Flash Avatar \cite{xiang2024flashavatar} or Splatting Avatar \cite{shao2024splattingavatar}. Unlike \shortname, these monocular methods do not generalize well to novel views and expressions. Moreover, they require three orders of magnitude more real data to create a single avatar. \shortname overcomes this by leveraging a synthetic prior during few shot avatar inversion, achieving high-quality results.

\section{Method}
\label{sec:method}

This section describes \shortname, how we train the synthetic prior to generate drivable 3D Gaussian head avatars, and how we use it for few-shot head avatar reconstruction.

\subsection{Preliminaries}

We represent a base 3D mesh as \mbox{$\mathbf{S} = \bar{\mathbf{S}} + \boldsymbol{\delta} \mathbf{B}_{id} + \boldsymbol{\gamma} \mathbf{B}_{expr}$}, where $\bar{\mathbf{S}}$ is the average shape, $\mathbf{B}_{(id, expr)}$ are the bases for identity and expression of a 3DMM, and $\boldsymbol{\delta}$, $\boldsymbol{\gamma}$ denote the corresponding coefficients. Additionally, we use linear blend skinning (LBS) for head rotation around the neck with pose corrective offsets, and to rotate the eyeballs.

The head avatar is rendered via 3D Gaussian splatting, using the CUDA implementation of 3DGS~\cite{kerbl2023gaussian}. The rasterizer is defined as $\mathcal{R}(\pmb{G}, \mathbf{K}) \rightarrow \splat$, for a camera $\mathbf{K}$ and a set of $n$ 3D Gaussians \mbox{$\pmb{G} \in \mathbb{R}^{n \times (11 + 16 \times 3)} := \{\pmb{\phi},\pmb{\theta}, \pmb{\sigma},\pmb{\alpha}, \pmb{h}\}$}, with position $\position \in \mathbb{R}^{n\times3}$, rotation $\rotation \in \mathbb{R}^{n\times3\times3}$, scale $\scale \in \mathbb{R}^{n\times3}$, opacity $\opacity \in \mathbb{R}^n$, and the (third-degree) spherical harmonics parameters $\sphericalharmonics \in \mathbb{R}^{n\times16\times3}$, where $n$ is the number of Gaussians.
See Kerbl \etal~\cite{kerbl2023gaussian} for more details.

\subsection{Gaussian Prior Model}
\label{sec:neuraltex}


Our prior is modeled as a generative convolutional network with additional lightweight regressors that output Gaussian 2D maps, i.e. multichannel parameter textures. To sample a flexible number of Gaussian primitives, UV positions and features are bilinearly interpolated from intermediary feature maps, before decoding the standard Gaussian attributes that are rendered using $\mathcal{R}(\cdot)$. 
The architecture of the prior learned by \shortname is illustrated in Fig.~\ref{fig:overview}.

\paragraph{Drivable VQ-VAE.}
Our network has an encoder-decoder architecture based on the VQ-VAE \cite{van2017neural}. 
We follow the approach of Esser \etal \cite{esser2020taming}, and use a transformer operating in a quantized latent codebook space to better model long-range dependencies between encoded patches in images.
The input to the encoder consists of an RGB texture map $\texturemap \in \texturemapdim$, an XYZ vertex position map $\vertexmap = \mathcal{R}_{uv}(\boldsymbol{\delta}\mathbf{B}_{id}) \in \vertexmapdim$ representing the rasterized positions of the neutral mesh, and an expression map $\expressionmap = \mathcal{R}_{uv}(\boldsymbol{\gamma} \mathbf{B}_{expr}) \in \expressionmapdim$ denoting rasterized expression offsets from the neutral mesh, where $\mathcal{R}_{uv}(\cdot)$ denotes UV space rasterization.
The encoder network consists of two parallel branches, one for identity and one for expression. 
This way we explicitly disentangle static components, such as face shape and appearance, from dynamic ones, such as wrinkles, and self-shadowing using two separate latent spaces.
We denote them as $\identityencoder(\texturemap, \vertexmap) \rightarrow \identitycode$, where $\identitycode \in \identitycodedim$ is the identity code and $\expressionencoder(\expressionmap) \rightarrow \expressioncode$ with $\expressioncode \in \expressioncodedim$ representing the expression code.
The identity and expression latents undergo element-wise quantization $\quantization(\cdot)$.
For simplicity, we omit the subscript and let $\latentcode \in \{\identitycode, \expressioncode\}$ denote identity and expression latent codes, with spatial codes $z_{ij} \in \mathbb{R}^{n}$, which we quantize by:
\begin{equation}
    \quantization(\latentcode) := \left( \underset{z_k \in \mathcal{Z}}{\arg \min} \, \| z_{ij} - z_k \| \right),
\end{equation}
with a learned discrete codebook $\mathcal{Z} = \{z_k\}_{k=1}^{K}$, with $z_k \in \mathbb{R}^{n}$.
The quantized latent codes are fed into the decoder, which is implemented as three output branches: a feature map decoder, $\featuredecoder(\quantization(\identitycode), \quantization(\expressioncode)) \rightarrow \outfeaturemap \in \outfeaturemapdim$ with $\featuredim$-dimensional feature vectors per texel; an identity map decoder, $\identitydecoder(\quantization(\identitycode)) \rightarrow \{\outtexturemap, \outvertexmap\}$; and an expression decoder, $\expressiondecoder(\quantization(\expressioncode)) \rightarrow \outexpressionmap$. Given the output vertex position and expression maps, $\outvertexmap$ and $\outexpressionmap$, the positions of the Gaussian primitives are then computed as $\position = \outvertexmap + \outexpressionmap$.

\paragraph{Gaussian Primitives Regression.}
\label{sec:primregress}
A common limitation of using CNNs to regress Gaussian maps is the fixed output resolution, which ties the number of primitives to the output dimensions.
This restriction can significantly limit the quality of the reconstructed avatar (see Table~\ref{tab:arch_ablation}). To address this issue, we use a part-based densification mechanism.
Similar to Kirschstein \etal~\cite{kirschstein2024gghead}, we use bilinear sampling, \( \sampling(\cdot, u, v) \) to sample the output of the decoders at UV-positions $(u, v)$.
As different head regions $r \in \{face, hair\}$ have varying requirements for the density of Gaussian primitives, we bilinearly sample separate parameter maps for the face and scalp region, rather than a single joint map.
Thus, per-part map sampling acts as adaptive primitive densification for the individual regions to improve visual quality (Table \ref{tab:arch_ablation}).
%

We define the primitive positions in the 3DMM space using only shape and expression. Global rotation, translation, and linear blend skinning (LBS) are factored out and applied to the primitives just before splatting to place them in the correct world space.
We compute initial \textit{per-part Gaussian parameters} for our primitives. Note that we do not use a fixed canonical space \cite{kirschstein2024gghead, zielonka2024gem, giebenhain2024npga}, as our initialization is derived from predicted position maps. We first obtain positions by sampling $\position_r = \mathcal{B}(\position, u_{r}, v_{r})$, for $r \in \{face, hair\}$. Next, for each $\position_r$, we compute nearest neighbor distance and initialize scale as $\pmb{\sigma}_r = \min_{j \neq i} \| \position_{r_i} - \position_{r_j} \|_2$. Initial opacity is set to $\pmb{\alpha} = 0.7$. Finally, the per-part rotations are computed as $\rotation_r = \begin{bmatrix} \frac{\mathbf{T}}{\|\mathbf{T}\|} & \frac{\mathbf{B}}{\|\mathbf{B}\|} & \frac{\mathbf{N}}{\|\mathbf{N}\|} \end{bmatrix}  \in \mathbb{R}^{h \times w \times 3 \times 3}$, based on the image space gradient:
\begin{align}
    \mathbf{T} = \frac{\partial \position_r}{\partial u}, \quad \mathbf{B} = \frac{\partial \position_r}{\partial v}, \quad \mathbf{N} = \mathbf{T} \times \mathbf{B}.
\end{align}
Following common practice \cite{giebenhain2024npga, zielonka2024gem, Zielonka2025Drivable3D, xu2023gaussianheadavatar, zheng2024headgap, kirschstein2024gghead, saito2024rgca}, we predict a neural corrective field for all Gaussian parameters.
For this, we use the regressed feature map $\outfeaturemap$, sampling $\sampledfeatures_r = \sampling(\outfeaturemap, u_{r}, v_{r})$, and lightweight regressors composed of four stacked convolutional blocks with skip connections.
Per region, we define two regressors:
\begin{align}
    \colorregressor(\sampledfeatures_r) & \rightarrow \sphericalharmonics_r  \in \mathbb{R}^{h \times w \times 16 \times 3}, \\
    \gaussianregressor(\sampledfeatures_r) & \rightarrow \{\delta \position_r, \delta \rotation_r, \delta \scale_r, \delta \opacity_r \},
\end{align}
where $\colorregressor$ regresses the spherical harmonics coefficients $\sphericalharmonics_r$, and $\gaussianregressor$ regresses additive parameter offsets
\mbox{$\Delta := \{ \delta \position_r, \delta \rotation_r, \delta \scale_r, \delta \opacity_r \}$} from the per-part Gaussian parameters.
Finally, we apply $\Delta$ to the primitives of the individual parts, concatenate them, and splat as $\mathcal{R}(\pmb{G}, \mathbf{K}) \rightarrow \splat$, where $\splat$ is the final rendered image and $\pmb{G}$ represents the combined Gaussian primitives.

\paragraph{Training Objectives.} 
\label{sec:losses}
We supervise the training of our model by minimizing the photometric loss:
\begin{align}
\mathcal{L}_{\text{color}} &= \alpha\mathcal{L}_{L1} + \beta \mathcal{L}_{\text{SSIM}} + \gamma\mathcal{L}_{\text{LPIPS}}
\label{eq:photo_loss}
\end{align}
between the pairs of input and output maps $\{\texturemap, \outtexturemap\}$, $\{\vertexmap, \outvertexmap\}$, and $\{\expressionmap, \outexpressionmap\}$, and between the pairs of target images and the final splatted images $\{\image, \splat\}$.

Additionally, the position maps $\outvertexmap$ and expression maps $\outexpressionmap$ are supervised by $\mathcal{L}_{\text{geom}} = \delta\mathcal{L}_{L1}$. 
The final loss is defined as $\mathcal{L} = \mathcal{L}_{\text{color}} + \mathcal{L}_{\text{geom}}$.
Moreover, we apply $L_2$ regularization on position, scale, opacity, and the FC ($l >= 1$) part of the spherical harmonics coefficients:
The final loss is defined as $\mathcal{L} = \mathcal{L}_{photo} + \mathcal{L}_{reg}$.
We train our network end-to-end using 8 GPUs Nvidia H100 with batch size 16 (2 per GPU).
We optimize the network for 500K iterations with the Adam optimizer \cite{kingma2014adam} with lr=$1.3e^{-5}$ and multi-step scheduler which decays the learning rate every milestone by gamma=0.66.

\subsection{Few-shot Avatar Reconstruction}
\label{sec:inversion}
\begin{figure}[t]
    \centering
    \setlength{\unitlength}{0.1\columnwidth}
    \begin{picture}(10, 2.3)
    \put(0, 0){\includegraphics[width=1.0\columnwidth]{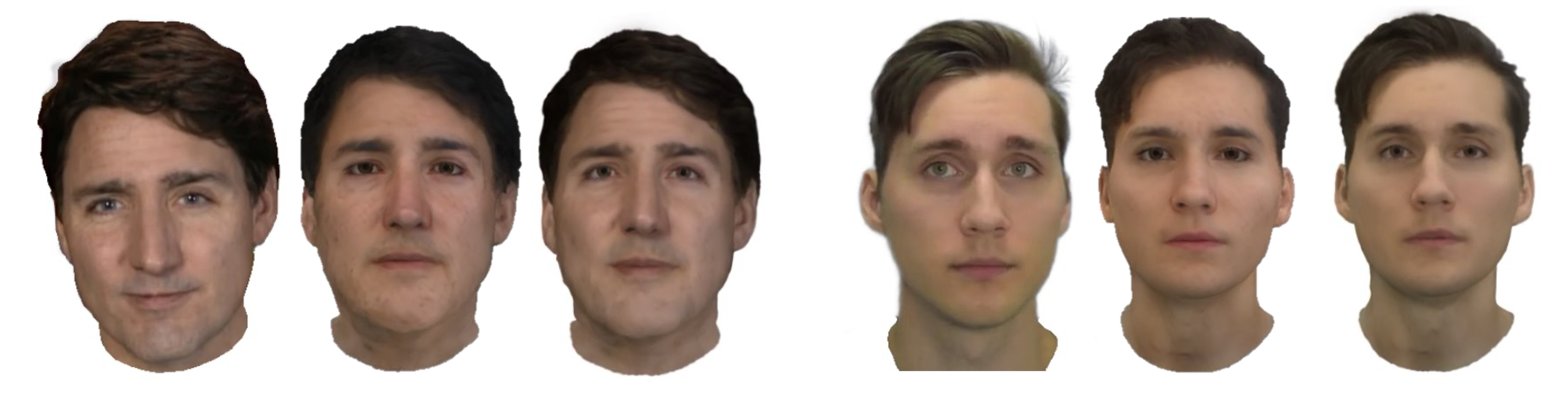}}
    \put(0.4, -0.2){Source}
    \put(2.2, -0.2){Prior}
    \put(3.7, -0.2){Final}
    \put(5.6, -0.2){Source}
    \put(7.3, -0.2){Prior}
    \put(8.8, -0.2){Final}
    \end{picture}
    \caption{
    Result of the pivotal tuning before (Prior) and after fine-tuning the model decoders and regressors (Final).
    }
    \label{fig:prior_fine_tune}
\end{figure}

To bridge the gap between in-the-wild and synthetic avatars, we carefully designed a two-stage inversion process based on pivotal fine-tuning \cite{pivotaltuning}.
First, we optimize the encoder $\identityencoder$ while keeping the rest of the network fixed such that we recover $\identitycode$.
Note that $\pmb{E}_{expr}$ remains unchanged as it should model independent expressions.
Once $\pmb{E}_{id}$ is fine-tuned, we fix its predicted identity latent code $\identitycode$, we fine-tune the decoders $\{\featuredecoder, \identitydecoder, \expressiondecoder\}$ and the regressors $\{\colorregressor, \gaussianregressor\}$ for the hair and face regions (Fig. \ref{fig:prior_fine_tune}).
To make the problem tractable, we employ a few heuristics to aid the optimization.
These include early stopping with a warmup phase and an exponential moving average on the loss to determine the stopping criteria.
Additionally, we scale the number of optimization steps based on the number of training frames, using a constant factor of 10 to increase the likelihood that each sample is seen at least once.
As a training objective, in addition to our photometric term $\mathcal{L}_{\text{color}}$ (Eq.~\ref{eq:photo_loss}), we follow Lattas \etal~\cite{lattas2023fitme} and, based on ArcFace~\cite{deng2019arcface}, define two additional objectives: $\mathcal{L}_{\text{id}}$ and $\mathcal{L}_{\text{arc}}$. The final inversion loss is equal to $\mathcal{L} = \mathcal{L}_{color} + \mathcal{L}_{arc} + \mathcal{L}_{id}$.
For a number of views, up to 20, the optimization takes less than \textbf{10} minutes on a single Nvidia H100 which is comparable to INSTA \cite{Zielonka2022InstantVH}.
The training time increases with the number of frames as we scale the iterations accordingly. 

\subsection{Synthetic Dataset}
\label{sec:syndata}

\begin{figure}[hb!]
    \centering
    \includegraphics[width=1.0\linewidth]{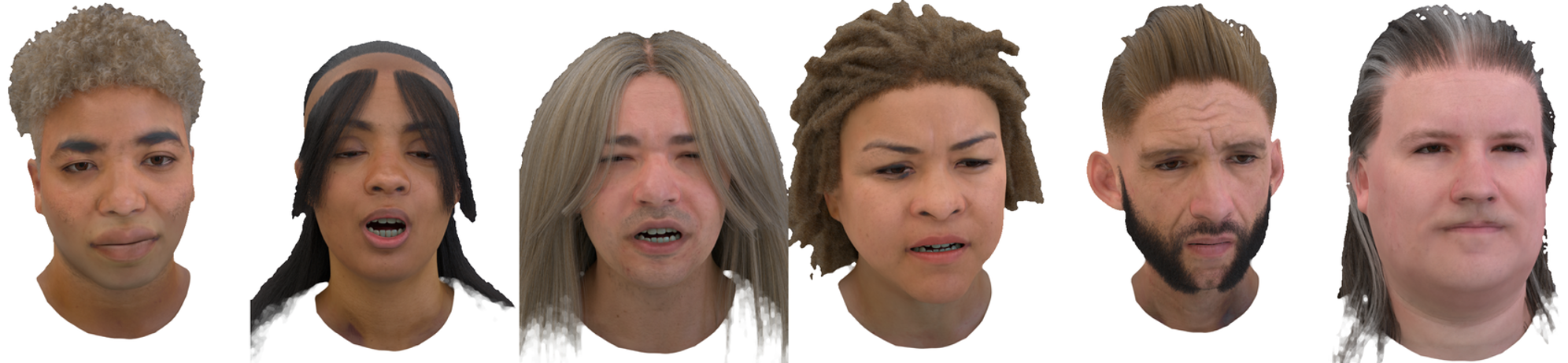}
    \caption{Random samples of our synthetic dataset show a diverse range of identities, expressions, and hairstyles that would be challenging to capture in an in-house studio with real subjects.}
    \label{fig:syn_dataset_1}
\end{figure}

Our dataset consists of approximately 2,000 unique identities, which we render with resolution $768\times512$ using Blender (Cycles) following Wood \etal~\cite{wood2021fake}, see Fig.~\ref{fig:syn_dataset_1}.
We positioned fourteen cameras in front of the subject and an additional fourteen cameras sampled from the upper hemisphere, centered on the scene.
We randomly assign assets such as hairstyles and beards to these avatars.
Additionally, we utilize high-quality face textures which are randomly distributed among the samples.
By combining different shapes and appearances, we augment the set of identities, following practices in synthetic data \cite{wood2021fake} and 3D face reconstruction \cite{lattas2023fitme, papantoniou2023relightify, dib2024mosar}.
To incorporate tracked expressions from multi-view setups, we propagate them to the avatars during sequence rendering.
We additionally compute a hair proxy from strands by voxelizing and fitting it to the scalp region; we apply the same approach for beards.
Using a neutral mesh and its hair proxy, we backproject the images onto the texture map.
During test time, we use a 3DMM regressor and the input images to extract a texture, which is then used as an initialization for our method.
In total, our dataset comprises 14 million images.

\begin{figure*}[t]
    \centering
    \setlength{\unitlength}{0.1\textwidth}
    \begin{picture}(10, 4.158)
    \put(0, 0){\includegraphics[width=1.0\linewidth]{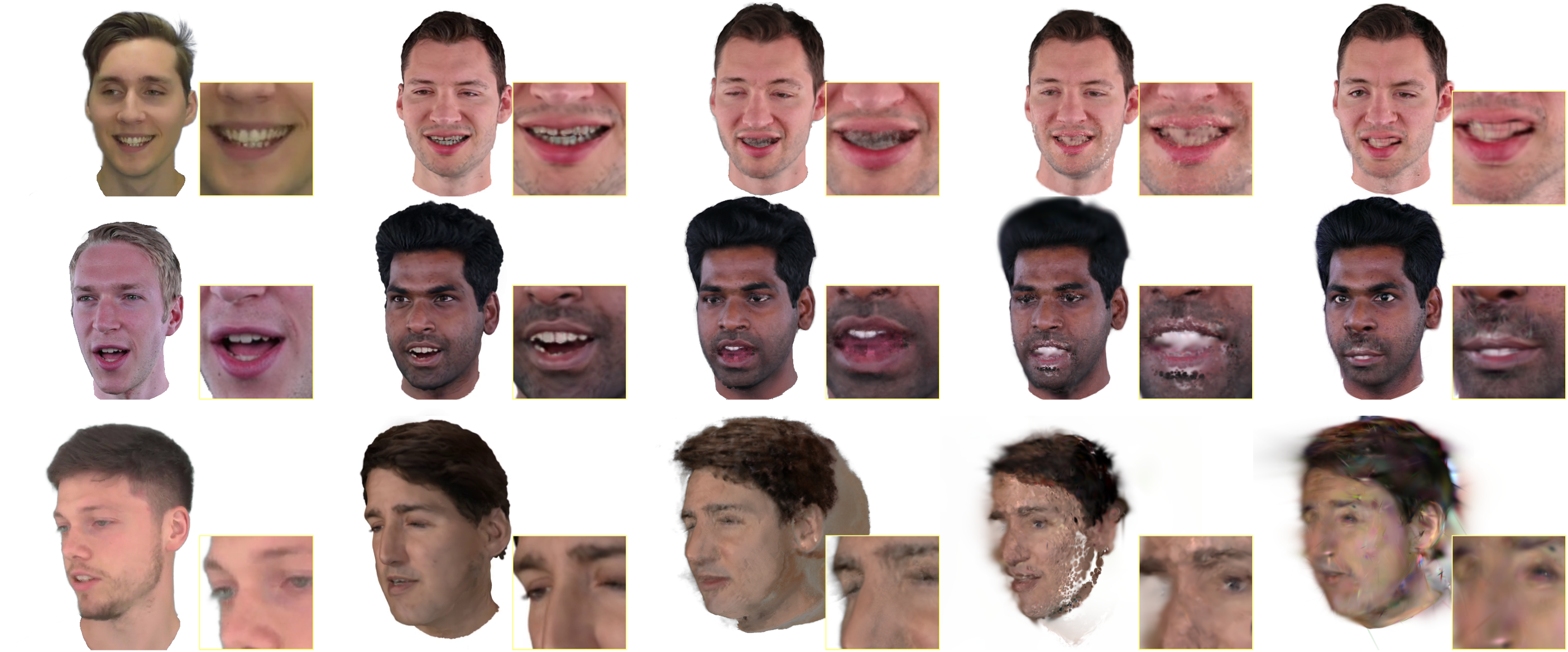}}
    \put(1.0, -0.3){Source}
    \put(3.0, -0.3){Ours}
    \put(4.8, -0.3){INSTA~\cite{Zielonka2022InstantVH}}
    \put(6.8, -0.3){SA \cite{shao2024splattingavatar}}
    \put(8.8, -0.3){FA \cite{xiang2024flashavatar}}
    \end{picture}
    \vspace{0.1cm}
    \caption{Cross-reenactment comparison of \shortname inversion using only \textbf{3} views to state-of-the-art (SOTA) methods: \textbf{INSTA} \cite{Zielonka2022InstantVH}, Flash Avatar (\textbf{FA}) \cite{xiang2024flashavatar}, and Splatting Avatar (\textbf{SA}) \cite{shao2024splattingavatar}, each of which was trained on an average of \textbf{3000} frames. It is evident that without a strong prior, these methods fail to generalize to novel expressions and views. Inversion input images are in the supplemental materials.}
    \label{fig:transfer_1}
\end{figure*}

\begin{figure}[t]
    \centering
    \includegraphics[width=0.95\linewidth]{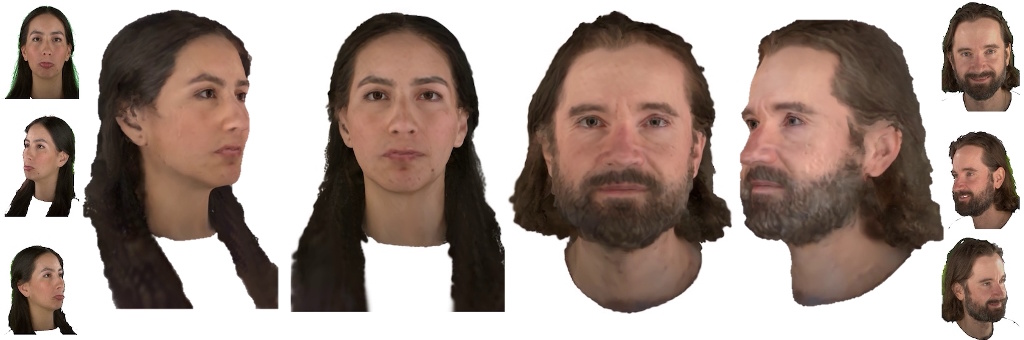}
    \caption{Novel view evaluation of long hair and beard inversion using only three input images demonstrates the strong generalization capability of SynShot.}
    \label{fig:hair_beard_row}
\end{figure}

\begin{figure}[t]
    \centering
    \includegraphics[width=1.0\columnwidth]{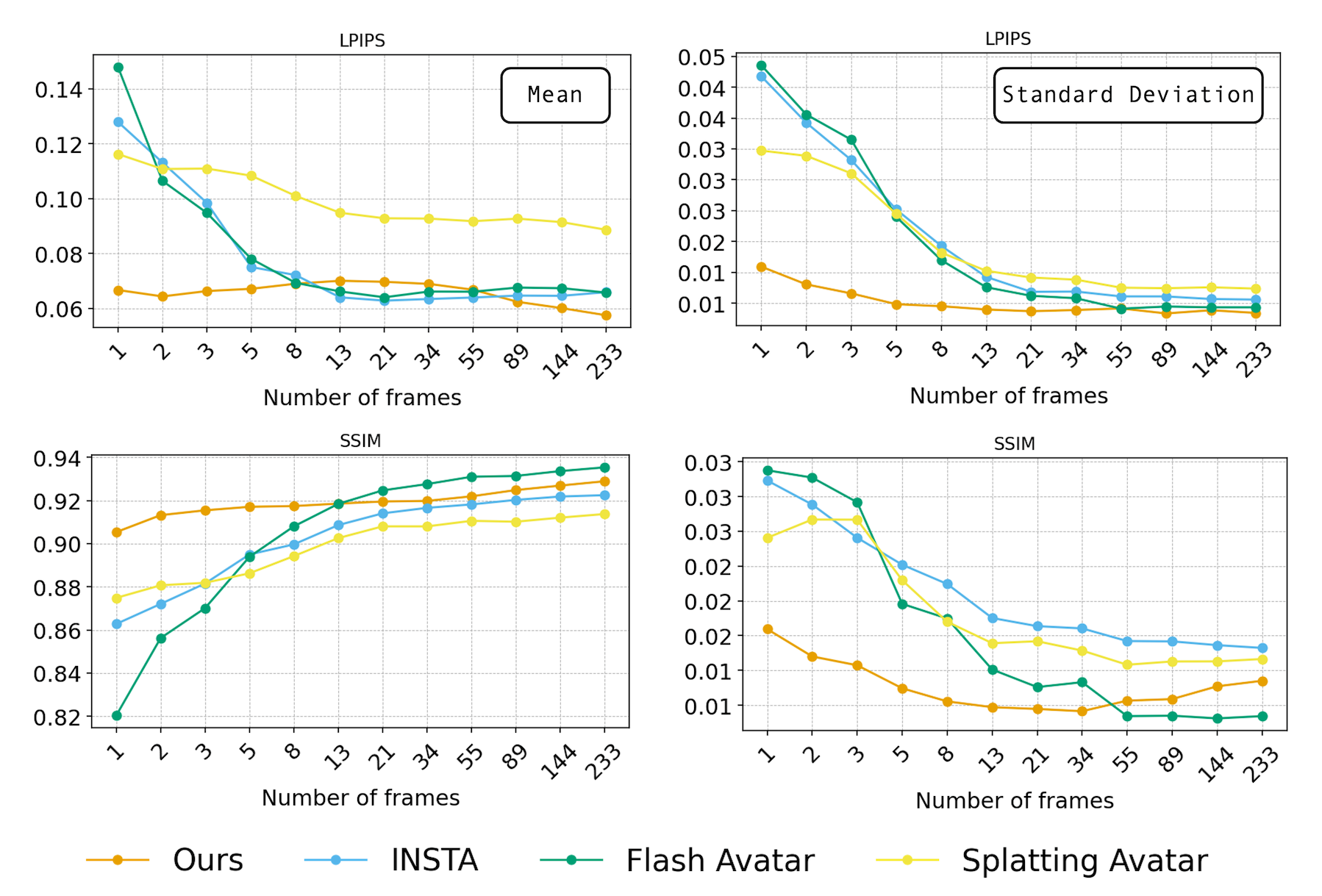}
    \caption{We evaluated the reconstruction error with respect to the number of frames using LPIPS and SSIM metrics. For each frame count, we report the average error (left) and standard deviation (right) over 600 frames across 11 subjects, highlighting the importance of our synthetic prior.}
    \label{fig:n_view_ablation}
\end{figure}

\begin{figure*}[h!]
    \centering
    \setlength{\unitlength}{0.1\textwidth}
    \begin{picture}(10, 3.739)
    \put(0, 0){\includegraphics[width=1.0\textwidth]{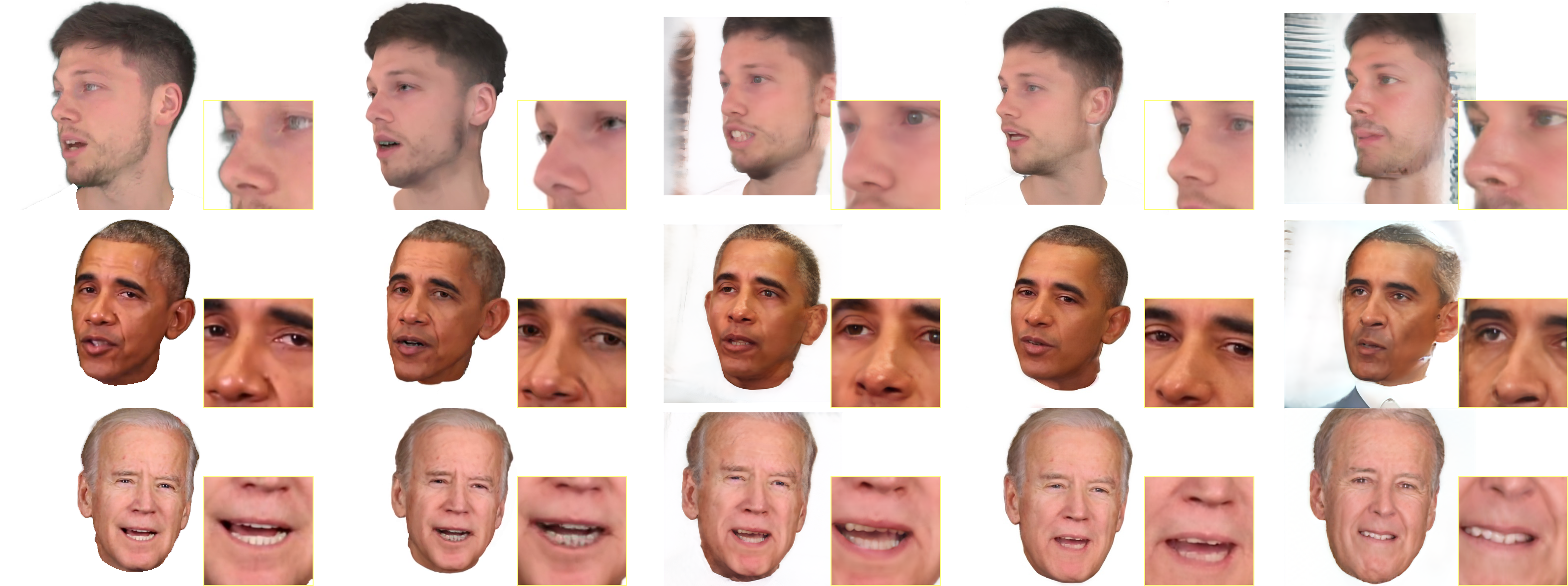}}
    \put(1.0, -0.3){Ground Truth}
    \put(3.0, -0.3){Ours}
    \put(4.5, -0.3){InvertAvatar \cite{invertavatar}}
    \put(6.8, -0.3){Portrait4D \cite{deng2024portrait4d}}
    \put(8.8, -0.3){Next3D \cite{sun2023next3d}}
    \end{picture}
    \vspace{0.1cm}
    \caption{The GAN-based self-reenactment comparison again shows that SynShot better captures identity and synthesizes novel views, proving its usefulness as a synthetic prior and for pivotal fine-tuning in inversion. LPIPS scores: Ours (\textbf{0.0236}), InvertAvatar (0.0962), Portrait4D (0.0843), and Next3D (0.2274). Inversion input images can be found in the supplemental materials.}
    \label{fig:gen_self_reanc}
\end{figure*}

\begin{figure*}[ht!]
    \centering
    \setlength{\unitlength}{0.1\textwidth}
    \begin{picture}(10, 3.975)
    \put(0, 0){\includegraphics[width=1.0\textwidth]{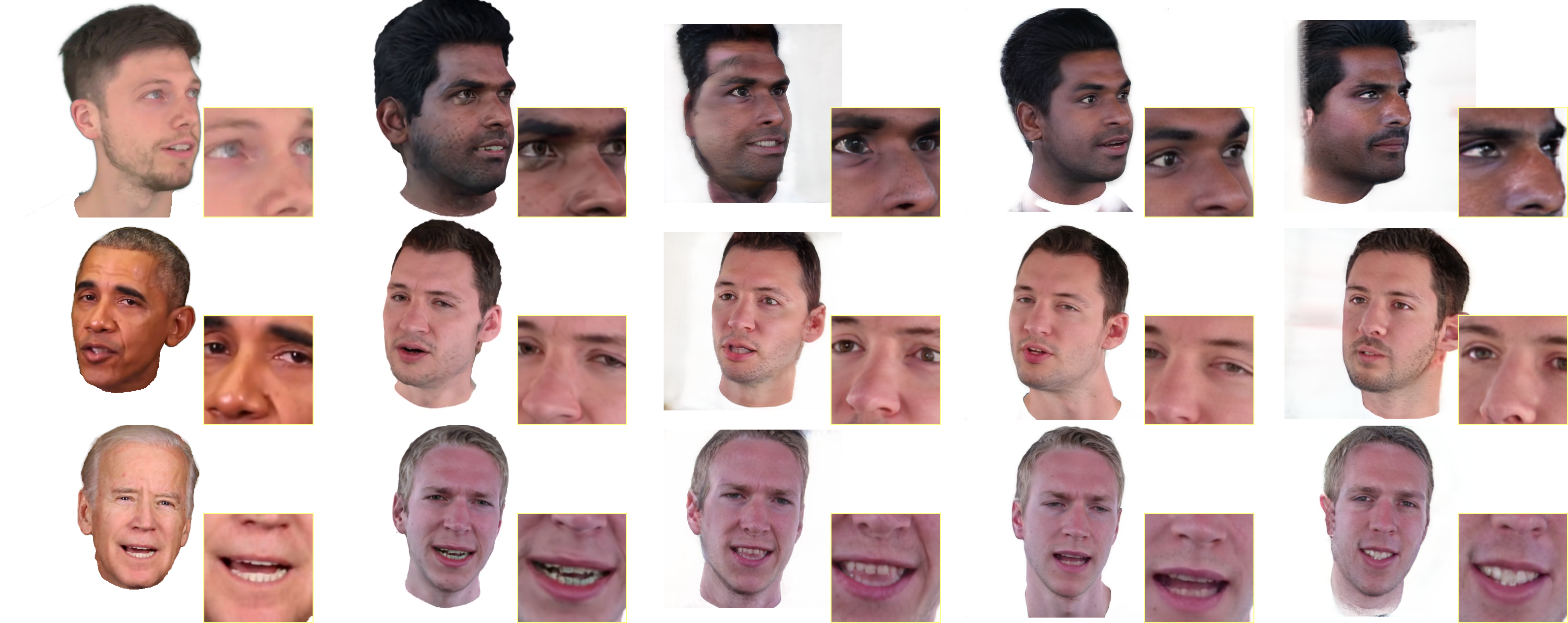}}
    \put(1.0, -0.3){Source}
    \put(3.0, -0.3){Ours}
    \put(4.5, -0.3){InvertAvatar \cite{invertavatar}}
    \put(6.8, -0.3){Portrait4D \cite{deng2024portrait4d}}
    \put(8.8, -0.3){Next3D \cite{sun2023next3d}}
    \end{picture}
    \vspace{0.1cm}
    \caption{The GAN-based cross-reenactment comparison shows that our method better reconstructs the target subject's appearance (identity) and remains faithful to the source subject's head poses and expressions, whereas the other methods suffer from artifacts.
    }
    \label{fig:gen_cross_reanac}
\end{figure*}

\section{Results}
\label{sec:results}
We compare \shortname to two different types of methods, state-of-the-art personalized monocular methods, and inversion-based general methods.
The personalized monocular methods are controlled by FLAME \cite{Li2017flame} meshes and include INSTA \cite{Zielonka2022InstantVH}, Flash Avatar \cite{xiang2024flashavatar}, and SplattingAvatar \cite{shao2024splattingavatar}.
For monocular methods, we used an ensemble of four datasets \cite{grassal2022neural, zielonka2024gem, zheng2022imavatar, Gafni2021nerface} processed using the face tracker from Zielonka \etal \cite{Zielonka2022TowardsMR}.
SplattingAvatar follows the approach of Zheng \etal \cite{zheng2022imavatar} and uses the monocular 3D face regressor DECA \cite{Feng2021Deca} for tracking.
In our experiments, we adopted a similar approach, employing an in-house regressor, similar to DECA, to estimate 3DMM expression and pose parameters.
While these methods produce photorealistic avatars, they struggle with generalization to novel views and poses (see Figure \ref{fig:transfer_1}).
For inversion-based methods, we compare PanoHead \cite{panohead2023}, HeadNeRF \cite{hong2021headnerf}, and MofaNeRF \cite{zhuang2022mofanerf}. We also compare to concurrent works including Portrait4D \cite{deng2024portrait4d}, Next3D \cite{sun2023next3d}, and InvertAvatar \cite{invertavatar} (Figures \ref{fig:gen_self_reanc} and \ref{fig:gen_cross_reanac}).
We use three images for all inversion experiments see supp. material.
Figure \ref{fig:hair_beard_row} presents a novel view evaluation of challenging long hair and beard inversion, demonstrating the generalization capabilities of \shortname. 

\paragraph{Evaluation.} To measure the performance of \shortname without introducing bias, we selected training frames from \(\{F_n\}_{n=1}^{16} = \{1, \dots, 987\}\), where \(F_n\) denotes the Fibonacci sequence.
For all experiments, we use progressive farthest point sampling \cite{qi2017pointnet} in the 3DMM expression space to select a specified number of frames from the training set. The self-reenactment sequences were evaluated using LPIPS and SSIM on the last 600 frames from the INSTA dataset \cite{Zielonka2022InstantVH}.

\paragraph{Monocular Avatar Self-Reenactment.}
Our combined dataset consists of eleven monocular sequences ($512\times512$ resolution), many of which are in-the-wild videos with very limited head motion, resulting in a low error as the test sequences closely resemble the training data, leaving limited room to assess diversity. To address this and accurately measure the effective error, we trained each method on a varying number of frames, corresponding to frames used in our inversion pipeline. The reconstruction error is evaluated on 600 test frames. Figure \ref{fig:n_view_ablation} demonstrates the effectiveness of our inversion, particularly with up to 233 training frames. Due to the lack of a strong prior, monocular methods fail in low training frame regimes, and, even with larger training datasets, they do not perform well and produce artifacts. Please note that to benefit from an increased number of input frames, i.e., to ground the avatar reconstruction more on the input than the synthetic prior, it requires an increased number of optimization iterations during pivotal fine-tuning. The number of iteration steps affects the metrics, causing LPIPS to vary non-monotonically.

\paragraph{Monocular Avatar Cross-Reenactment.}
We would like to emphasize the importance of evaluating cross-reenactment, which often reveals issues with generalization and overfitting; however, these aspects are frequently underemphasized, as evaluation sequences are commonly not sufficiently challenging.
For instance, Figure~\ref{fig:n_view_ablation} indicates that 13 frames may be sufficient for monocular methods to perform well on the test set. Despite achieving high-quality results, most monocular methods \cite{grassal2022neural, Zielonka2022InstantVH, Gafni2021nerface, xiang2024flashavatar, shao2024splattingavatar} struggle with cross-reenactment involving novel expressions and views. In the supp. mat. we present a full evaluation. Without a strong prior, these methods frequently exhibit artifacts when driven by out-of-distribution sequences. In contrast, our method, leveraging only three images and a synthetic prior with effective shape-expression disentanglement, is able to invert an avatar that significantly outperforms state-of-the-art models trained on thousands of frames. Figure \ref{fig:transfer_1} demonstrates cross-reenactment, with the leftmost column serving as the source for expression and view. This shows that incorporating a strong prior enhances the visual quality.

\begin{table}[h]
    \centering
    \resizebox{1.0\linewidth}{!}{
        \begin{tabular}{lrrrr}
        \toprule
        Architecture & L1 $\downarrow$ & LPIPS $\downarrow$ & SSIM $\uparrow$ & PSNR $\uparrow$ \\
        \midrule
        $F=128$ & \textit{0.0356} &  \textbf{0.2686} & 0.8189 & \textit{20.1536} \\
        Tex. up-sampling & \textbf{0.0352} & \textit{0.2695} & \textbf{0.8196} & \textbf{20.1909} \\
        Single Layer & 0.0369 & 0.2702 & 0.8177 & 19.8871 \\
        $F=32$ & 0.0375 & 0.2732 & 0.8146 & 19.7002 \\
        w/o VQ & 0.0396 & 0.2747 & 0.8122 & 19.2861 \\
        $F=64$ & 0.0400 & 0.2765 & 0.8104 & 19.2731 \\
        No Sampling & 0.0403 & 0.2853 & 0.8158 & 19.9787 \\
        $256\times256$ & 0.0365 & 0.2865 & \textit{0.8194} & 20.4010 \\
        \bottomrule
        \end{tabular}
    }
    \caption{We evaluated various configurations of our VQ-VAE. Each configuration uses the final textures of $512 \times 512$, unless stated otherwise. As our final model ($F=128$) we selected the one which produces sharpest results in terms of LPIPS.}
    \label{tab:arch_ablation}
\end{table}

\paragraph{GAN-based baselines.}
We compared SynShot to three animatable GAN-based methods. For our method and InvertAvatar \cite{invertavatar}, we used three input images, whereas Portrait4D \cite{deng2024portrait4d} and Next3D \cite{sun2023next3d} are single-shot. Figure \ref{fig:gen_self_reanc} presents qualitative self-reenactment results, with additional quantitative LPIPS scores: Ours (\textbf{0.0236}), InvertAvatar (0.0962), Portrait4D (0.0843), and Next3D (0.2274). Both results show that SynShot significantly outperforms the baselines. Moreover, Figure \ref{fig:gen_cross_reanac} presents expression transfer, where our method best captures the subject's identity and is more stable for novel views and expressions, whereas GAN-based methods tend to introduce artifacts in side views.

\paragraph{VQ-VAE Architecture Ablation.}
Table \ref{tab:arch_ablation} presents an ablation study of our VQ-VAE architecture.
Each model was evaluated on 50 test actors excluded from the training set.
Our best model, in terms of sharpness and quality, regresses a feature map $\outfeaturemap \in \outfeaturemapdim$, where $\featuredim = 128$, at a resolution of $512 \times 512$.
Regressing Gaussian primitives directly (\textit{No Sampling}) suffers from lack of quality.
Using a \textit{Single Layer} instead of two (for hair + face) results in a lower number of Gaussians, which also decreases the final quality.
A key feature of our network is densification through texture sampling.
In the (\textit{Tex. up-sampling}) experiment, we predict feature maps at $256 \times 256$ resolution compared to $512 \times 512$ and apply bilinear sampling to upscale the per-region sampled feature maps to $512 \times 512$.
This approach achieves results that are almost on par while saving VQ-VAE computation and memory.
Finally, using codebook quantization of latent space improves the final image quality (\textit{w/o VQ}).
\begin{table}[h]
    \centering
    \resizebox{1.0\linewidth}{!}{
        \begin{tabular}{lrrrr}
        \toprule
        Loss & L1 $\downarrow$ & LPIPS $\downarrow$ & SSIM $\uparrow$ & PSNR $\uparrow$ \\
        \midrule
        $\mathcal{L}_{photo} + \mathcal{L}_{VGG}$ + $\mathcal{L}_{Id} + \mathcal{L}_{ArcFeat}$ & 0.0229 & 0.0776 & 0.9073 & 23.7474 \\
        $\mathcal{L}_{photo} + \mathcal{L}_{VGG}$ & 0.0244 & 0.0839 & 0.9058 & 23.1191 \\
        $\mathcal{L}_{photo} + \mathcal{L}_{VGG}$ + $\mathcal{L}_{Id}$ & 0.0246 & 0.0848 & 0.9048 & 23.1949 \\
        $\mathcal{L}_{photo} = \mathcal{L}_{L1}$ + $\mathcal{L}_{SSIM}$ & 0.0217 & 0.0904 & 0.9094 & 23.7331 \\
        \bottomrule
        \end{tabular}
    }
    \caption{Ablation for our inversion losses.}
    \label{tab:inversion_ablation}
\end{table}

\paragraph{Inversion Ablation.}
Our inversion pipeline consists of several losses that help bridge the gap between synthetic and real images.
This is an important step in our pipeline, as real subjects often have appearance and illumination conditions that differ significantly from our distribution.
To address this, we rely on pixel-wise losses and, most importantly, on perceptual losses, which have been shown to aid in effectively matching two distributions \cite{johnson2016perceptual, lattas2023fitme, zhang2018perceptual, buehler2024cafca, buehler2023preface}.
Table \ref{tab:inversion_ablation} shows the inversion reconstruction error using different combinations of losses.
As can be seen, using only $\mathcal{L}_{photo}$ is insufficient. The combination of $\mathcal{L}_{VGG}$, based on AlexNet \cite{krizhevsky2012imagenet}, $\mathcal{L}_{ID}$, and $\mathcal{L}_{ArcFeat}$ provides the best results.

\section{Discussion}
\label{sec:discussion}

While significantly outperforming monocular methods, \shortname has certain limitations that we identify.
A key challenge is bridging the domain gap between synthetic and real data.
There is considerable room for improvement in the generation of synthetic data.
For example, all our synthetic subjects share the same teeth geometry and texture.
As a consequence, teeth in our inverted head avatars often closely follow the prior and do not adapt easily.
Furthermore, our synthetic data lacks diverse expression-dependent wrinkles, affecting its overall visual quality. 
Additionally, our dataset was ray-traced with a single environment map, limiting generalization to varied lighting conditions.
%

\section{Conclusion}
\label{sec:conclusion}

We have proposed \shortname, a method for reconstructing a personalized 3D Gaussian head avatar from just a few images.
\shortname builds a generative head avatar purely from synthetic data and then utilizes this model as a prior in an inversion pipeline. 
This inversion pipeline follows a pivotal tuning strategy that successfully bridges the domain gap between the prior and the real input images.
We demonstrate that our personalized head avatars generalize better to unseen expressions and viewpoints than SOTA head avatars.
%
%
%
%

\paragraph{Acknowledgement}
The authors thank the International Max Planck Research School for Intelligent Systems (IMPRS-IS) for supporting WZ.
JT was supported by the ERC Starting Grant LeMo (101162081). We also would like to thank Daoye Wang for assisting with synthetic asset generation, Mark Murphy for his help with using Google infrastructure, and Menglei Chai for providing the hair proxy.

{
    \small
    \bibliographystyle{ieeenat_fullname}
    \bibliography{main}
}

\clearpage
\appendix
\twocolumn[{%
\renewcommand\twocolumn[1][]{#1}%
\begin{center}
    \textbf{\Large{\longname \\ -- Supplemental Document --}}
    \centering
    \vspace{1.1cm}
    \captionsetup{type=figure}
    \includegraphics[width=1.0\textwidth]{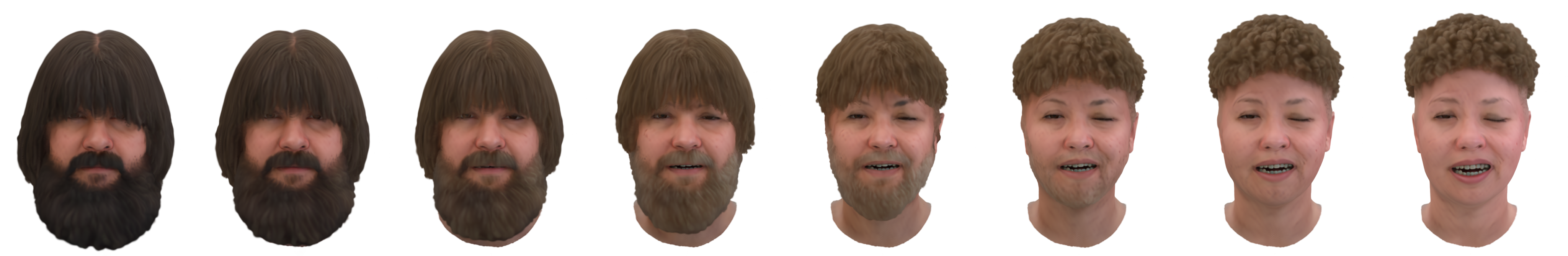}
    \vspace{0.3cm}
    \caption{Linearly interpolating $z_{\mathbf{q}_{\text{id}}}$ and  $z_{\mathbf{q}_{\text{expr}}}$ between the leftmost and rightmost avatars demonstrates that our latent manifold exhibits smooth transitions in both expression and identity.}
    \label{fig:supp-teaser}
\end{center}%
}]

\section{Appendix}

This supplementary material includes additional comparisons with monocular methods such as INSTA \cite{Zielonka2022InstantVH}, Flash Avatar (FA) \cite{xiang2024flashavatar}, and Splatting Avatar \cite{shao2024splattingavatar}, as well as comparisons with single-image-based reconstruction methods like PanoHead \cite{panohead2023}, MoFaNeRF \cite{zhuang2022mofanerf}, and HeadNeRF \cite{hong2021headnerf}. Additionally, we present inversions on a more diverse set of subjects, along with failure cases.

\begin{figure*}[ht!]
    \centering
    \setlength{\unitlength}{0.1\linewidth}
    \begin{picture}(10, 12.275)
    \put(0.7, 0){\includegraphics[width=0.85\linewidth]{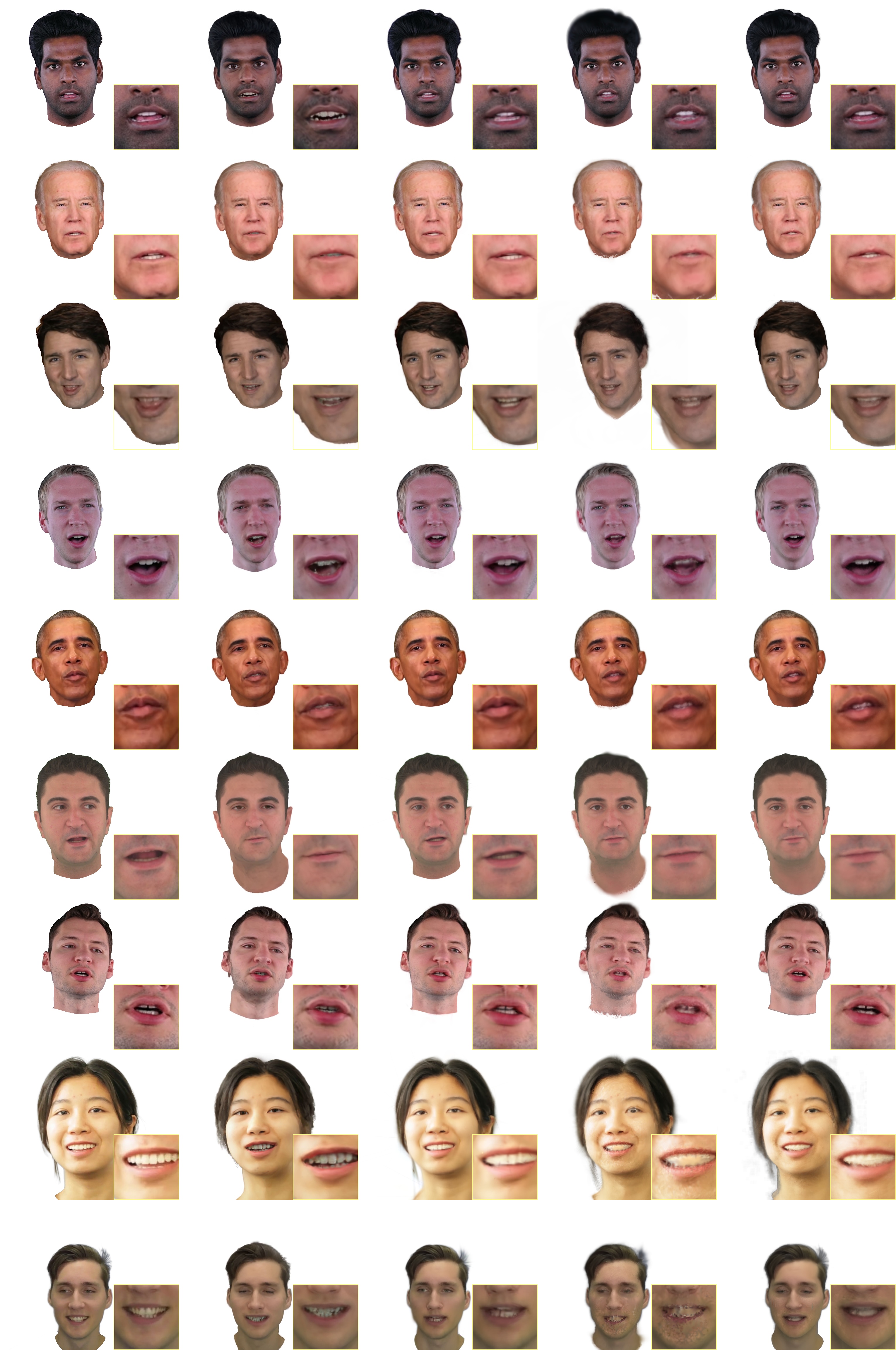}}
    \put(1.0, -0.3){Source}
    \put(3.0, -0.3){Ours}
    \put(4.8, -0.3){INSTA~\cite{Zielonka2022InstantVH}}
    \put(6.8, -0.3){SA \cite{shao2024splattingavatar}}
    \put(8.8, -0.3){FA \cite{xiang2024flashavatar}}
    \end{picture}
    \vspace{0.1cm}
    \caption{Test View Evaluation: When comparing the test views, which are very close to the training distribution, all baselines perform comparably well. Our method also achieves good results, despite the prior model being insufficiently refined in some cases (e.g., teeth).}
    \label{fig:test_view_mono}
\end{figure*}

\begin{figure*}[ht!]
    \centering
    \setlength{\unitlength}{0.1\linewidth}
    \begin{picture}(10, 11.4)
    \put(1.3, 0){\includegraphics[width=0.7\linewidth]{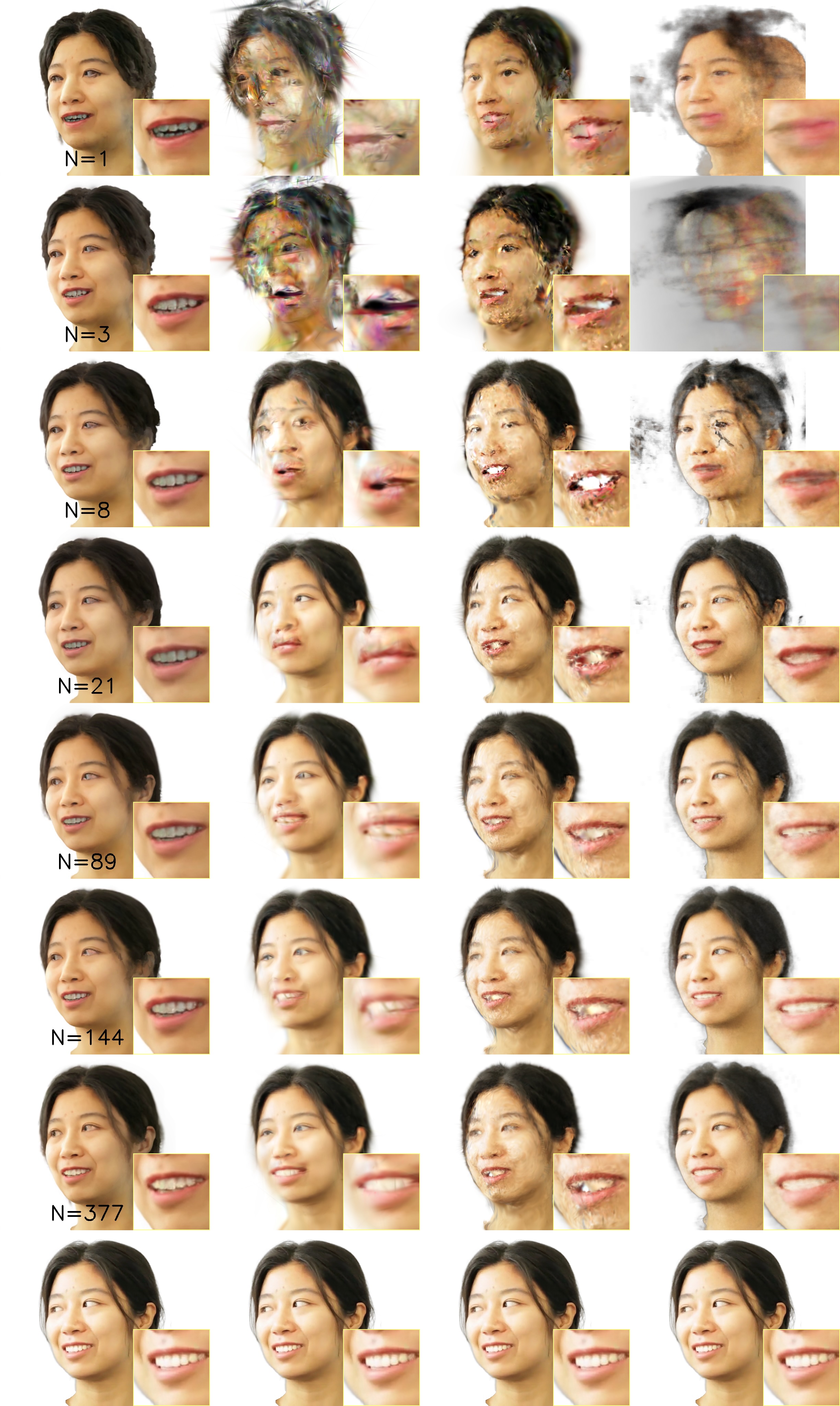}}
    \put(1.2, 0.1){\rotatebox{90}{Ground Truth}}
    \put(1.9, -0.3){Ours}
    \put(3.7, -0.3){FA \cite{xiang2024flashavatar}}
    \put(5.6, -0.3){SA \cite{shao2024splattingavatar}}
    \put(7.1, -0.3){INSTA \cite{Zielonka2022InstantVH}}
    \end{picture}
    \vspace{0.25cm}
    \caption{We trained each method on a different number of frames to demonstrate the importance of our prior model using test sequences. In this experiment, we progressively increased the number of training frames up to 377. The frames were sampled from the training set using Farthest Point Sampling defined on the 3DMM expression space. The comparison includes INSTA \cite{Zielonka2022InstantVH}, Flash Avatar (FA) \cite{xiang2024flashavatar}, and Splatting Avatar (SA) \cite{shao2024splattingavatar}.}
    \label{fig:n_views_baselines_yufeng}
\end{figure*}

\begin{figure*}[ht!]
    \centering
    \setlength{\unitlength}{0.1\linewidth}
    \begin{picture}(10, 11.4)
    \put(1.3, 0){\includegraphics[width=0.7\linewidth]{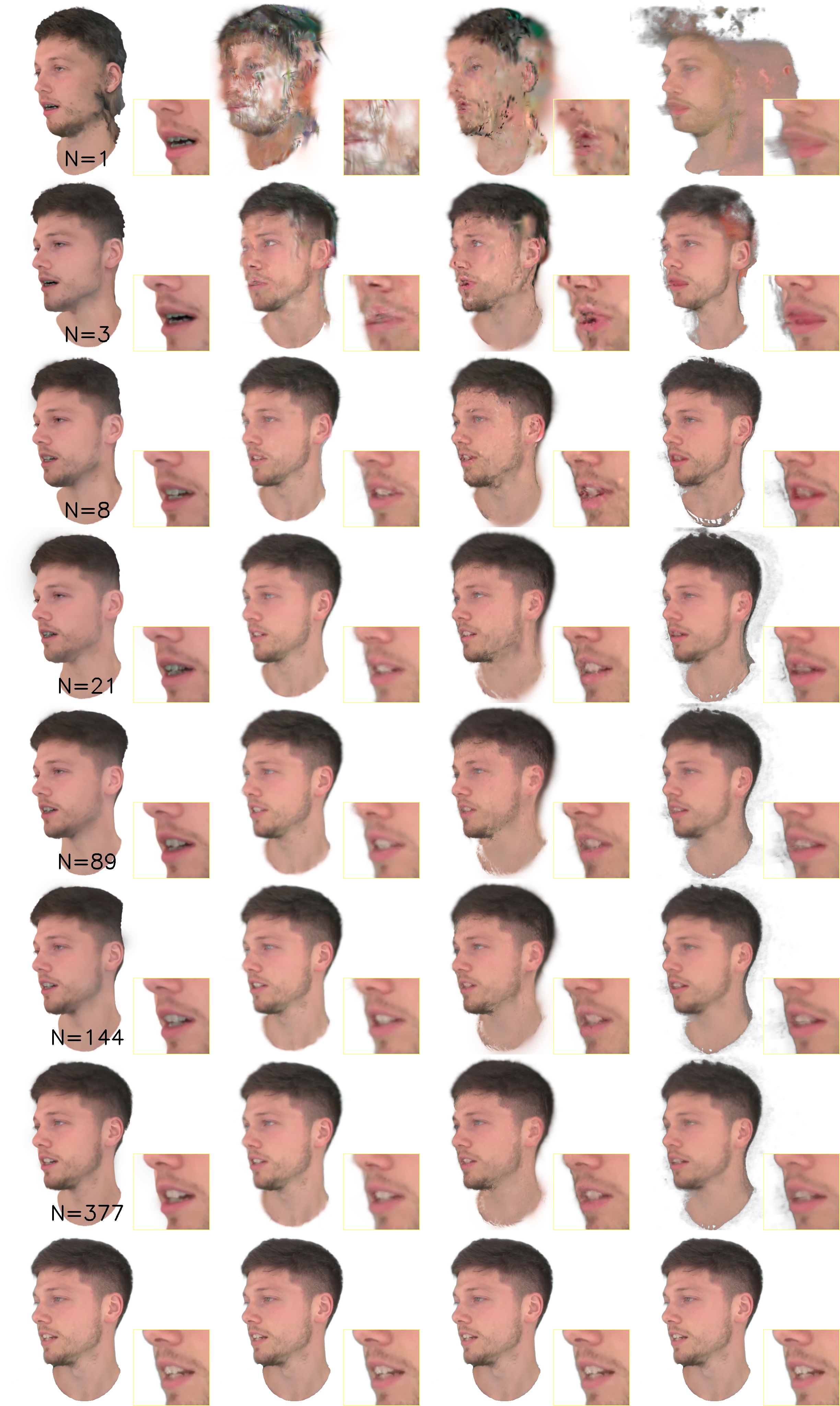}}
    \put(1.2, 0.1){\rotatebox{90}{Ground Truth}}
    \put(1.9, -0.3){Ours}
    \put(3.7, -0.3){FA \cite{xiang2024flashavatar}}
    \put(5.6, -0.3){SA \cite{shao2024splattingavatar}}
    \put(7.1, -0.3){INSTA \cite{Zielonka2022InstantVH}}
    \end{picture}
    \vspace{0.25cm}
    \caption{We trained each method on a different number of frames to demonstrate the importance of our prior model using test sequences. In this experiment, we progressively increased the number of training frames up to 377. The frames were sampled from the training set using Farthest Point Sampling defined on the 3DMM expression space. The comparison includes INSTA \cite{Zielonka2022InstantVH}, Flash Avatar (FA) \cite{xiang2024flashavatar}, and Splatting Avatar (SA) \cite{shao2024splattingavatar}.}
    \label{fig:n_views_baselines_person_0000}
\end{figure*}

\begin{figure*}[t!]
    \centering
    \includegraphics[width=1.0\linewidth]{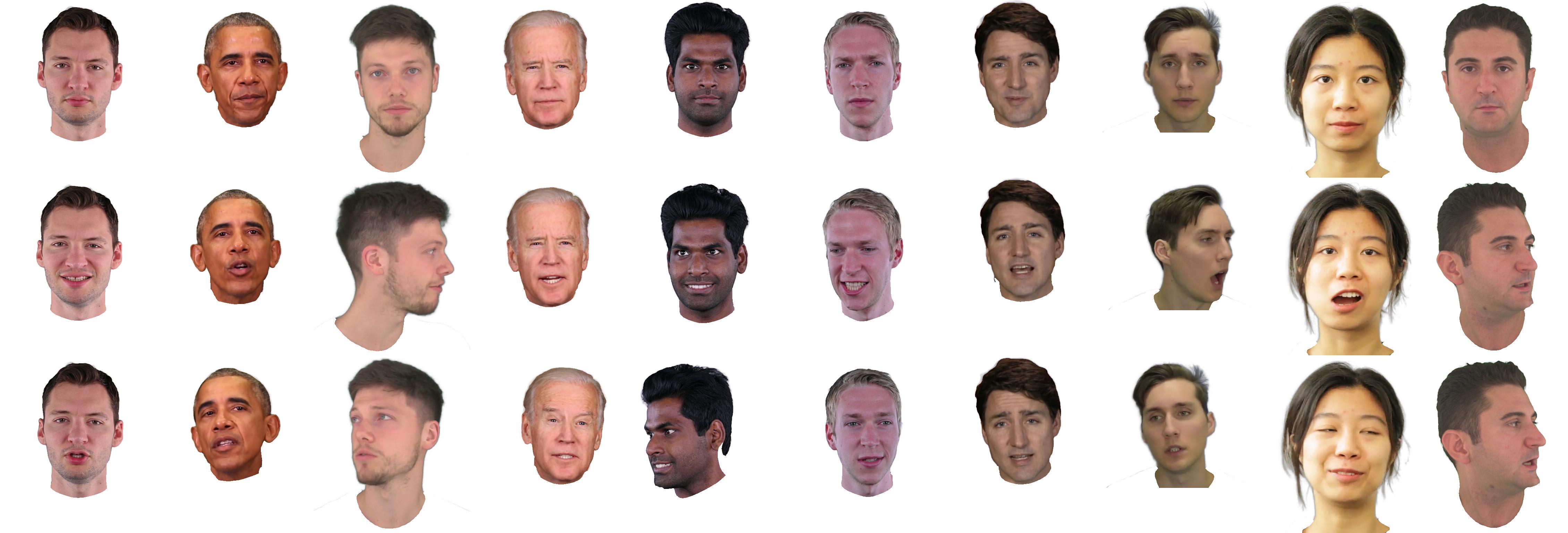}
    \caption{Unless otherwise stated, all experiments in this paper used three input images. Here, we present these images for each actor.}
    \label{fig:synshot_input}
\end{figure*}

All our inversion results used only three input images (Figure \ref{fig:synshot_input}) unless stated otherwise. Figure \ref{fig:test_view_mono} compares monocular baseline methods trained on the entire dataset with our inversion approach. Furthermore, we provide additional examples of cross-reenactment comparisons, demonstrating the advantages of our method compared to baselines trained on only 13 frames. Next, we present results with progressively varying numbers of training frames, illustrating how this influences the quality of reconstruction. 
Figures \ref{fig:n_views_baselines_yufeng} and \ref{fig:n_views_baselines_person_0000} highlight the importance of our synthetic prior. 

\begin{figure*}[ht!]
    \centering
    \vspace{-1.0cm}
    \begin{subfigure}[t]{0.48\textwidth}
        \centering
        \includegraphics[width=\linewidth]{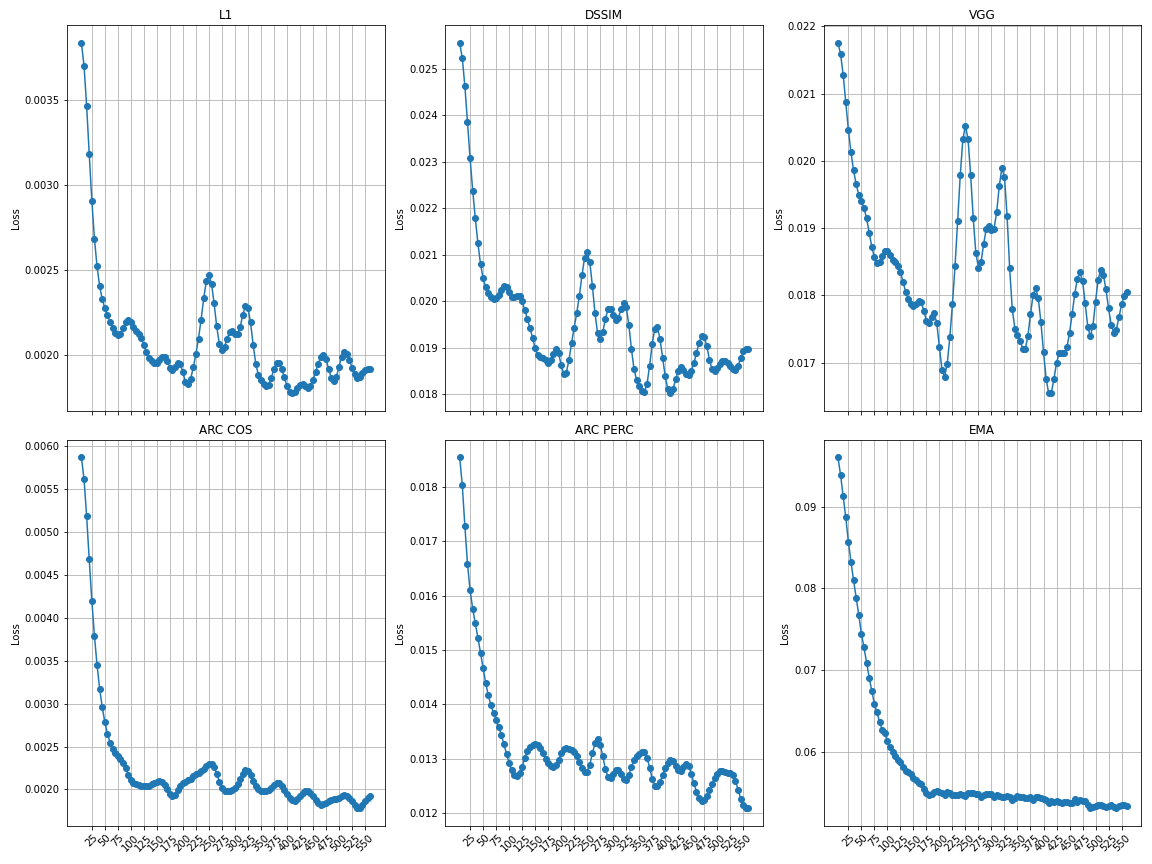}
        \caption{Our pivotal fine-tuning first stage: In this part, we optimize only the identity encoder to find the optimal projection of the input image onto our synthetic latent space.}
        \label{fig:latent_opt_diagram}
    \end{subfigure}
    \hfill
    \begin{subfigure}[t]{0.48\textwidth}
        \centering
        \includegraphics[width=\linewidth]{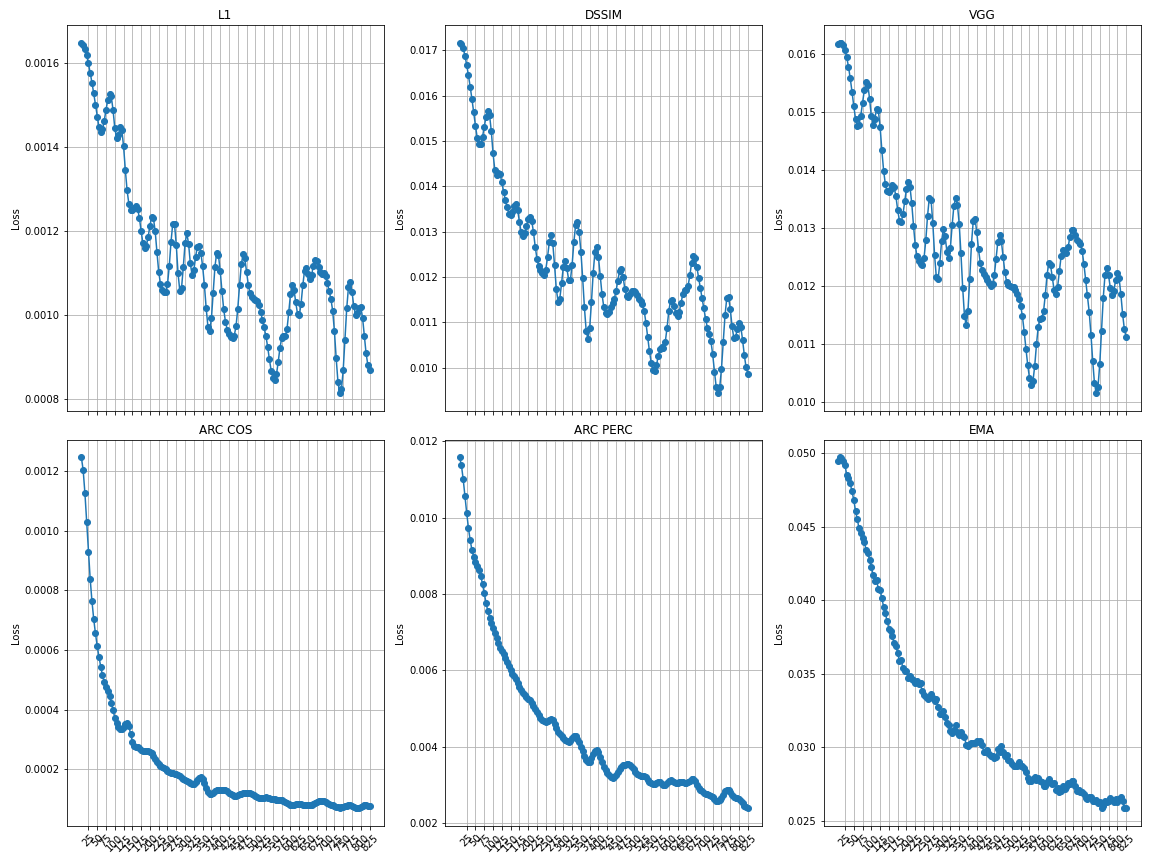}
        \caption{Our pivotal second stage of fine-tuning involves fixing the optimization latent code and focusing on optimizing the decoder to bridge the domain gap between the synthetic avatar and real subjects. During this phase, we typically address global illumination, identity texture, teeth color, and hair appearance by refining the decoders.}
        \label{fig:net_opt_diagram}
    \end{subfigure}
    \caption{An overview of the two pivotal fine-tuning stages. (a) The first stage optimizes the identity encoder. (b) The second stage optimizes the decoder to bridge the domain gap between synthetic avatars and real subjects.}
    \label{fig:combined_diagram}
\end{figure*}

\begin{figure}[t!]
    \centering
    \includegraphics[width=1.0\linewidth]{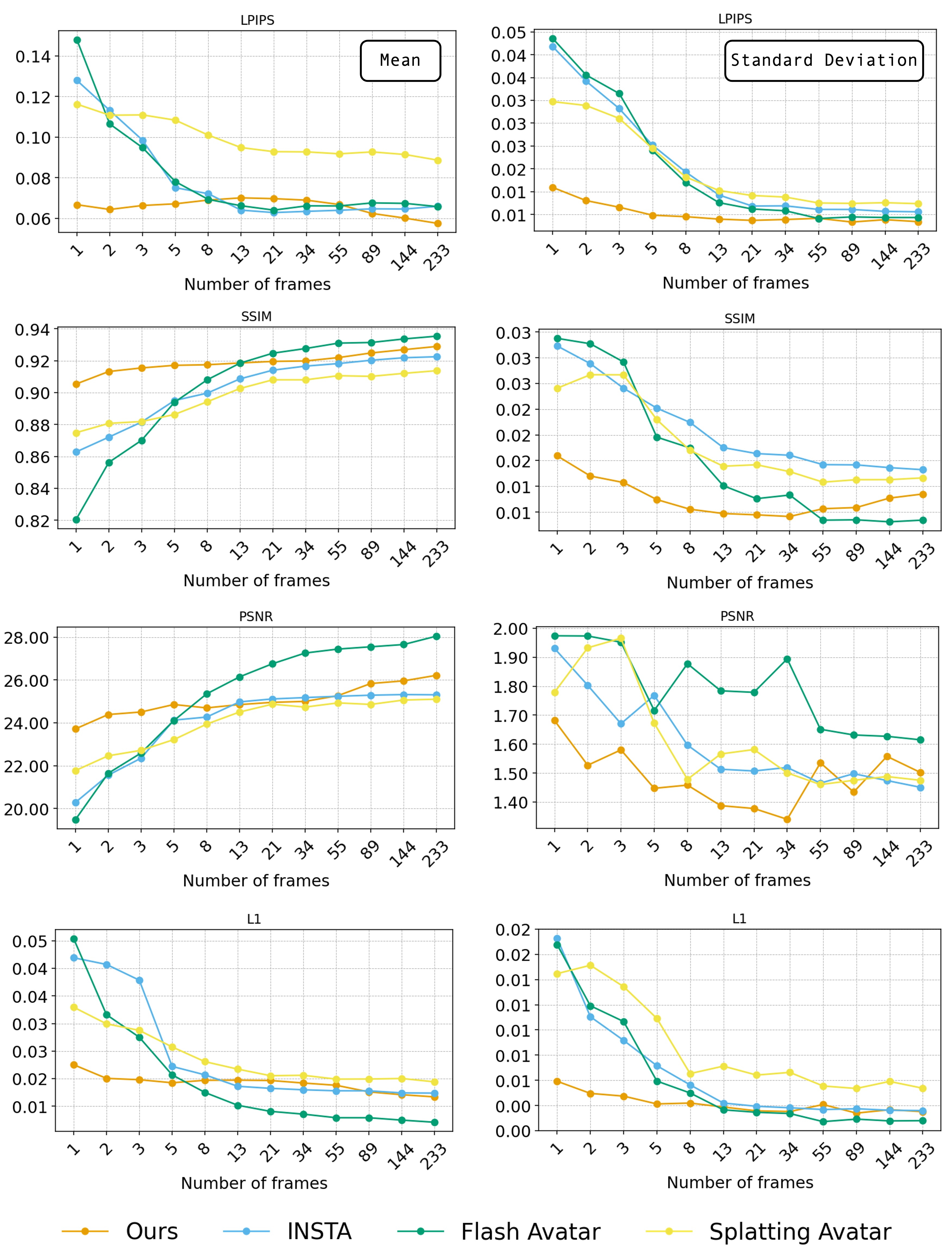}
    \caption{We evaluated the reconstruction error with respect to the number of frames using LPIPS, SSIM, L1, and PSNR metrics. For each frame count, we report the average error (left) and standard deviation (right) over 600 frames across 11 subjects.}
    \label{fig:plot_big}
\end{figure}

\begin{figure*}[ht!]
    \centering
    \setlength{\unitlength}{0.1\linewidth}
    \begin{picture}(10, 11)
    \put(0, 0){\includegraphics[width=1.0\linewidth]{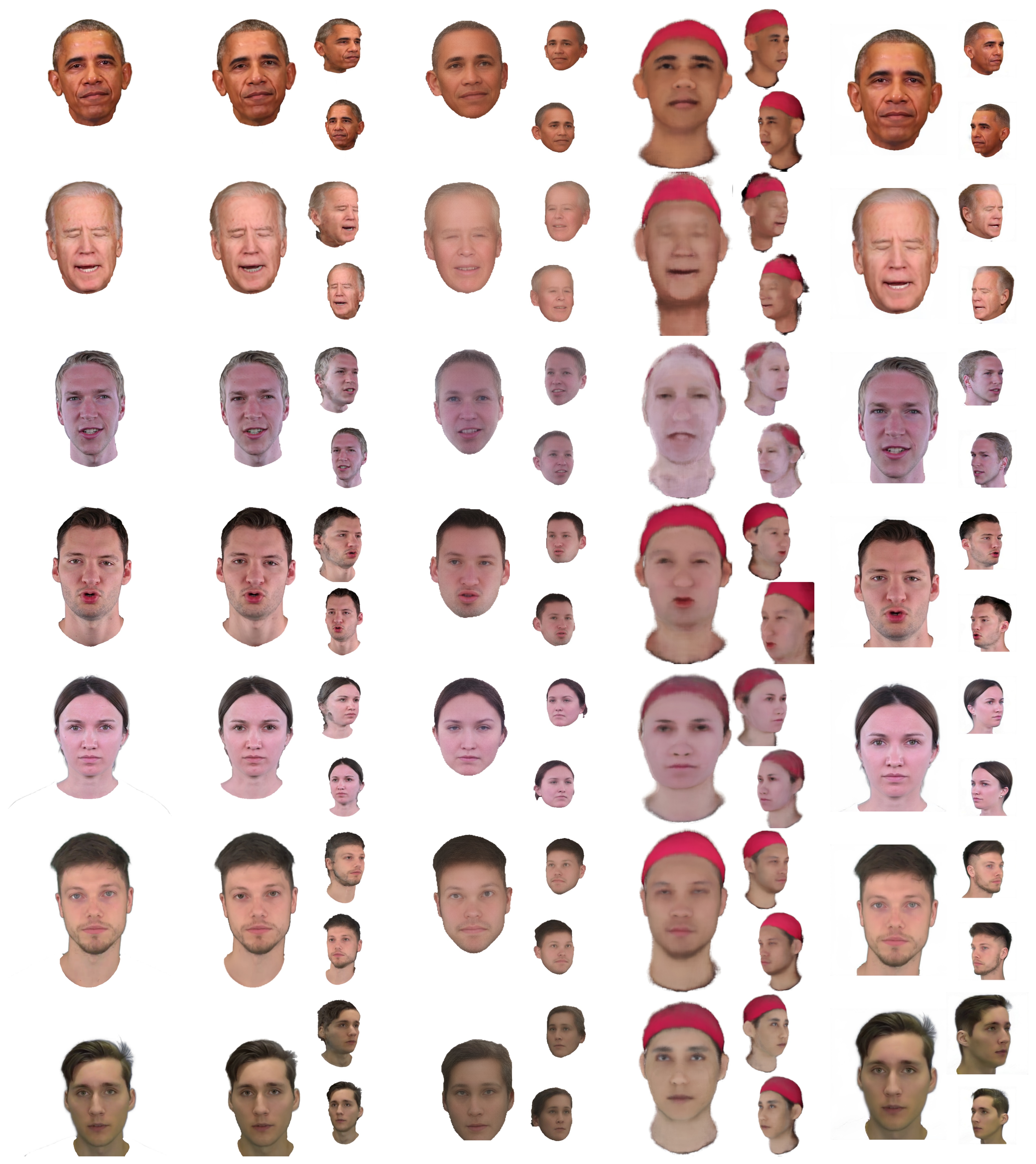}}
    \put(0.6, -0.2){Source}
    \put(2.6, -0.2){Ours}
    \put(4.2, -0.2){HeadNerf \cite{hong2021headnerf}}
    \put(6.4, -0.2){MofaNerf \cite{zhuang2022mofanerf}}
    \put(8.4, -0.2){PanoHead \cite{panohead2023}}
    \end{picture}
    \vspace{0.25cm}
    \caption{Additional results of single image inversion.}
    \label{fig:inversion_big}
\end{figure*}

\begin{figure*}[t!]
    \centering
    \includegraphics[width=1.0\linewidth]{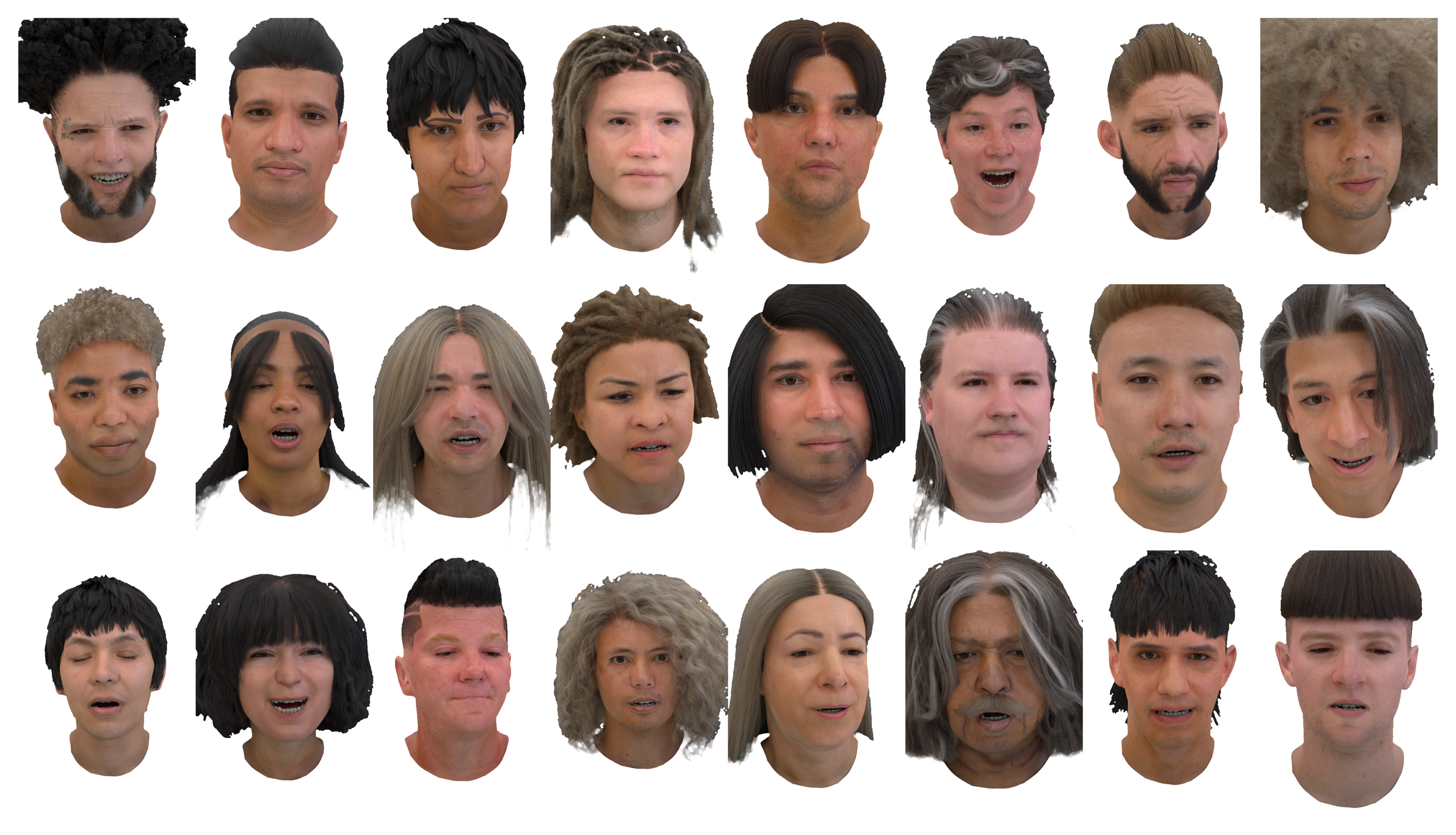}
    \caption{Random samples from our synthetic dataset, showcasing a diverse range of identities, expressions, and hairstyles that would be challenging to capture in an in-house studio with real subjects.}
    \label{fig:syn_dataset}
\end{figure*}

We include comparisons to single-image inversion methods in Figure \ref{fig:inversion_big}, and the losses diagrams for each stage in Figure \ref{fig:combined_diagram}. We also present additional samples from our synthetic dataset in Figure \ref{fig:syn_dataset}, as well as more interpolation steps for our identity $z_{\mathbf{q}_{\text{id}}}$ and expression $z_{\mathbf{q}_{\text{expr}}}$ latent spaces, shown in Figure \ref{fig:supp-teaser}. Finally, we complement the reconstruction error evaluation with additional metrics Figure \ref{fig:plot_big}.

\paragraph{Inversion Objectives} We depict the inversion optimization loss for one subject using three images as input. We show two stages of our pivotal fine-tuning: Figure \ref{fig:latent_opt_diagram} presents identity encoder optimization in the first stage, and Figure \ref{fig:net_opt_diagram} presents the second stage, where the decoding part of our pipeline is optimized. In this particular case, the optimization took around 5 minutes on a single Nvidia H100.

\begin{figure*}[ht!]
    \centering
    \vspace{-0.5cm}
    \setlength{\unitlength}{0.1\linewidth}
    \begin{picture}(15, 12.9)
    \put(0.4, 0){\includegraphics[width=0.95\linewidth]{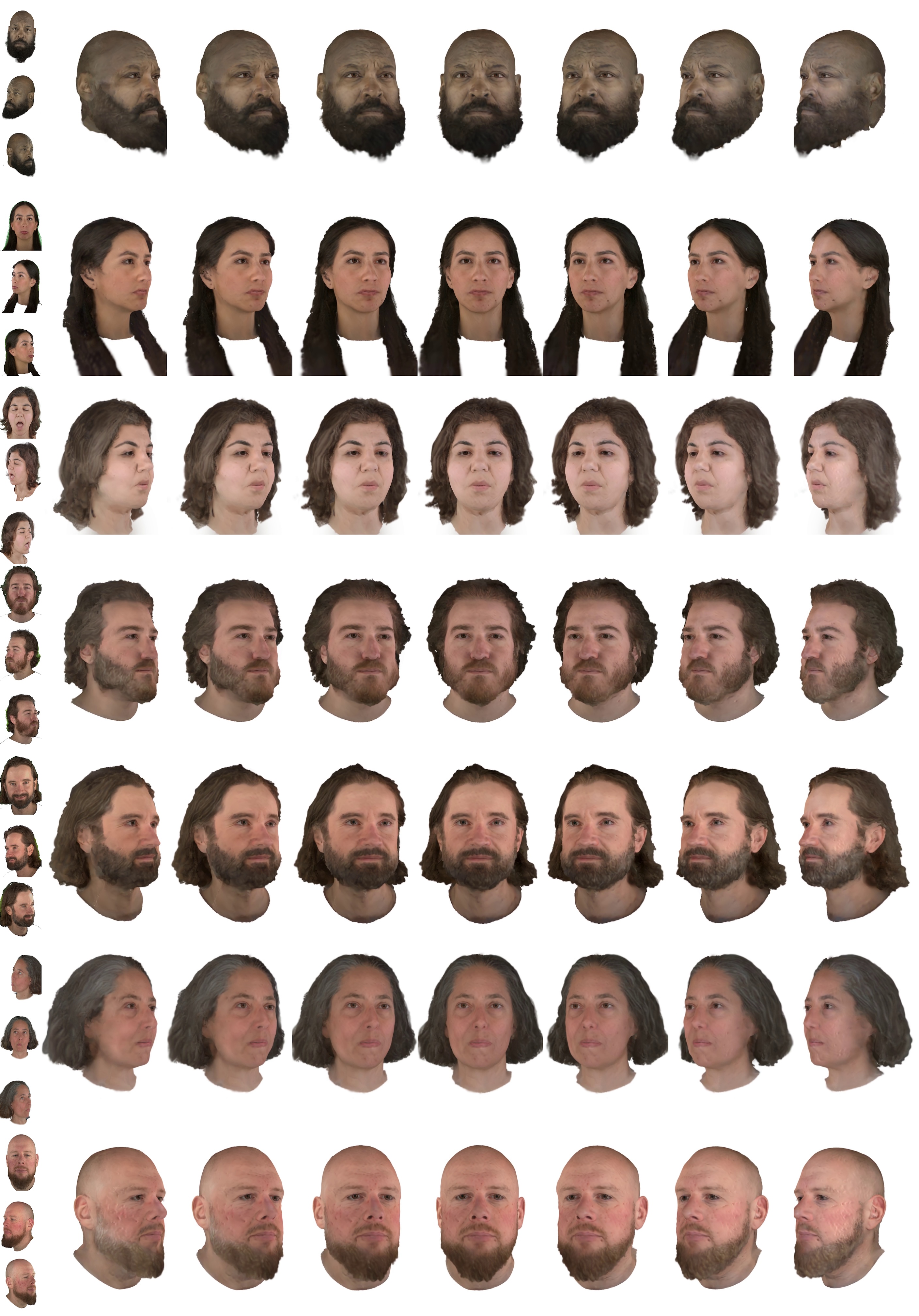}}
    \put(0.1, 0.45){\rotatebox{90}{Input Images}}
    \end{picture}
    \vspace{-0.5cm}
    \caption{Novel view evaluation of long hair and beard inversion using only three input images demonstrates the strong generalization capability of \shortname, which accurately models both long hair and beards.}
    \label{fig:inversion_beard_hair}
\end{figure*}

\begin{figure*}[ht!]
    \centering
    \setlength{\unitlength}{0.1\linewidth}
    \begin{picture}(15, 8)
    \put(0.4, 0){\includegraphics[width=0.95\linewidth]{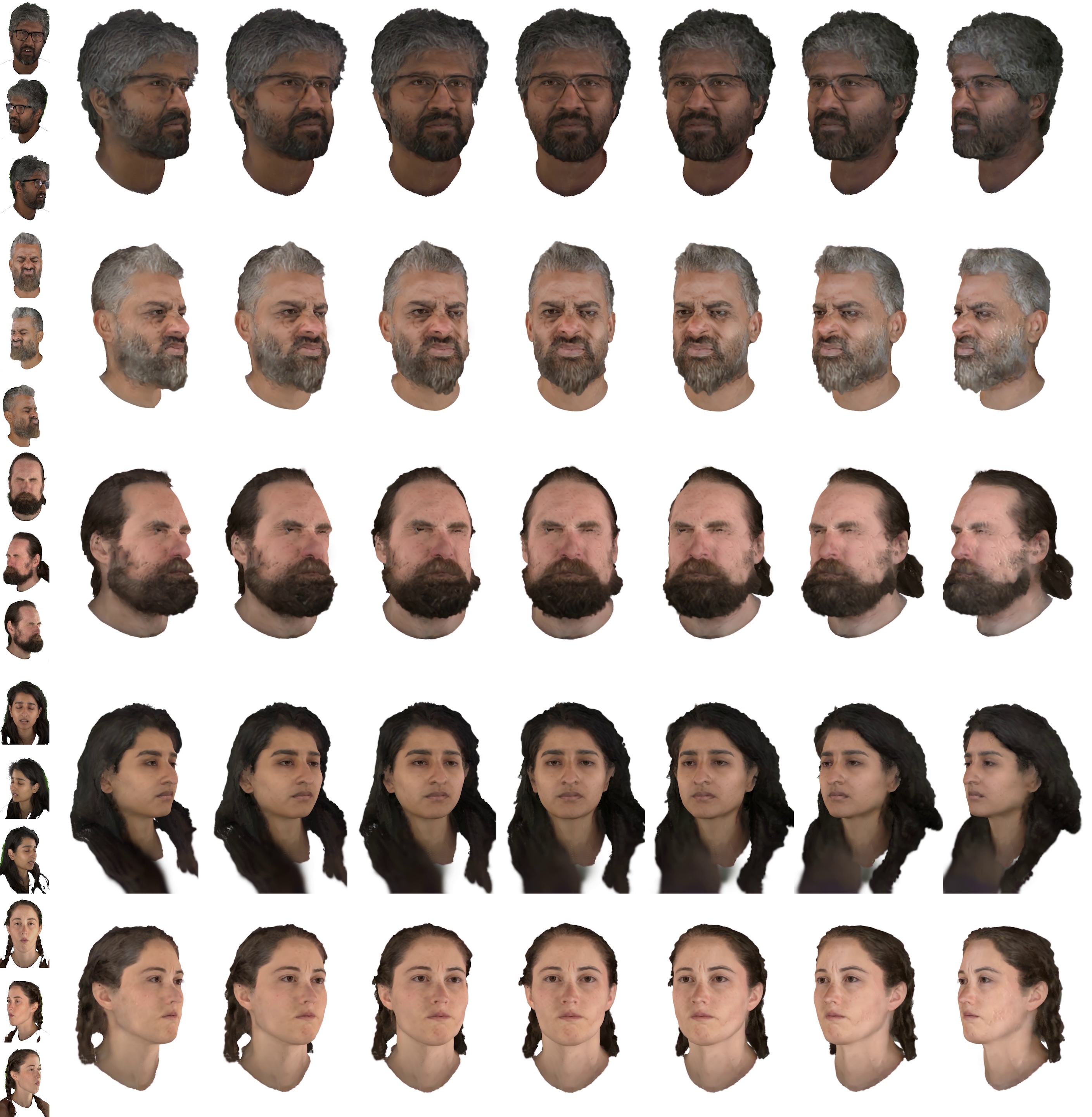}}
    \put(0.1, 0.45){\rotatebox{90}{Input Images}}
    \end{picture}
    \vspace{-0.1cm}
    \caption{Additionally, we present failure cases of our method, categorized into primary scenarios where \shortname may fail (from the top): (1) input images with facial accessories, such as glasses, which are absent from our synthetic dataset; (2) challenging inputs, such as squinting or closed eyes, which introduce artifacts in the final avatar; and (3) missing hairstyles in the dataset, leading to inversion errors for uncommon styles, further exacerbated by artifacts in hair segmentation.}
    \label{fig:failure_cases}
\end{figure*}

\paragraph{Additional Results} Figure \ref{fig:inversion_beard_hair} illustrates a challenging inversion for identities with long hair and beards, where \shortname successfully models these features using subjects from Preface \cite{buehler2023preface} dataset. Additionally, we present the failure cases of our method, categorized into the primary scenarios where \shortname may fail. As shown in Figure~\ref{fig:failure_cases}: 

\begin{itemize}
    \item \textbf{A)} Input images with facial accessories like glasses are not supported currently as they were not used in our synthetic dataset.
    \item \textbf{B)} Challenging input images, such as those with squinting eyes or closed eyes, can introduce artifacts in the final avatar due to difficulties in faithfully reproducing these details.
    \item \textbf{C)} Missing hairstyles in the synthetic dataset often result in errors during inversion, particularly for uncommon or complex hairstyles, further exacerbated by artifacts in hair segmentation.
\end{itemize}

\section{3D Gaussian Splatting Preliminaries}

3D Gaussian Splatting (3DGS)~\cite{kerbl2023gaussian} provides an alternative to Neural Radiance Field (NeRF)~\cite{mildenhall2020nerf} for reconstructing and rendering static multi-view scenes from novel perspectives.
Kerbl \etal~\cite{kerbl2023gaussian} represent the 3D space using scaled 3D Gaussians~\cite{Wang2019DifferentiableSS, Kopanas2021PointBasedNR}, defined by a 3D covariance matrix $\mathbf{\Sigma}$ and a mean $\mathbf{\mu}$:
\begin{equation}
\label{formula: gaussian's formula}
    G(\mathbf{x})=e^{-\frac{1}{2}(\mathbf{x-\mu})^T\mathbf{\Sigma}^{-1}(\mathbf{x-\mu})}.
\end{equation}
To render this representation, Zwicker \etal~\cite{Zwicker2001SurfaceS} project 3D Gaussians onto the image plane using the formula $\mathbf{\Sigma}^{\prime} = \mathbf{A}\mathbf{W}\mathbf{\Sigma} \mathbf{W}^T\mathbf{A}^T$, where $\mathbf{\Sigma}^{\prime}$ denotes the 2D covariance matrix. Here, $\mathbf{W}$ is the view transformation, and $\mathbf{A}$ is the projective transformation.
Rather than directly optimizing the covariance matrix $\mathbf{\Sigma}$, which must remain positive semidefinite, Kerbl \etal~\cite{kerbl2023gaussian} parameterize it in terms of scale $\mathbf{S}$ and rotation $\mathbf{R}$. This reformulation expresses the 3D Gaussian as a 3D ellipsoid: $\mathbf{\Sigma} = \mathbf{R}\mathbf{S}\mathbf{S}^T\mathbf{R}^T$.
Finally, 3DGS leverages the approach of Ramamoorthi \etal~\cite{Ramamoorthi2001AnER} to approximate the diffuse component of the BRDF~\cite{Goral1984ModelingTI} using spherical harmonics (SH) for modeling global illumination and view-dependent color.
Four SH bands are utilized, resulting in a 48-element vector.

\section{Broader Impact}
Our project centers on reconstructing highly detailed human face avatars from multiview videos, allowing for the extrapolation of expressions beyond those originally captured. While our technology is intended for constructive applications, such as enhancing telepresence and mixed reality experiences, we recognize the risks associated with misuse. To mitigate these risks, we advocate for progress in digital media forensics \cite{Rssler2019FaceForensicsLT, Rssler2018FaceForensicsAL} to support the detection of synthetic media. We also stress the importance of conducting research in this field with transparency and integrity.

\begin{figure*}[ht!]
    \centering
    \setlength{\unitlength}{0.1\textwidth}
    \begin{picture}(10, 10)
    \put(0, 0){\includegraphics[width=1.0\linewidth]{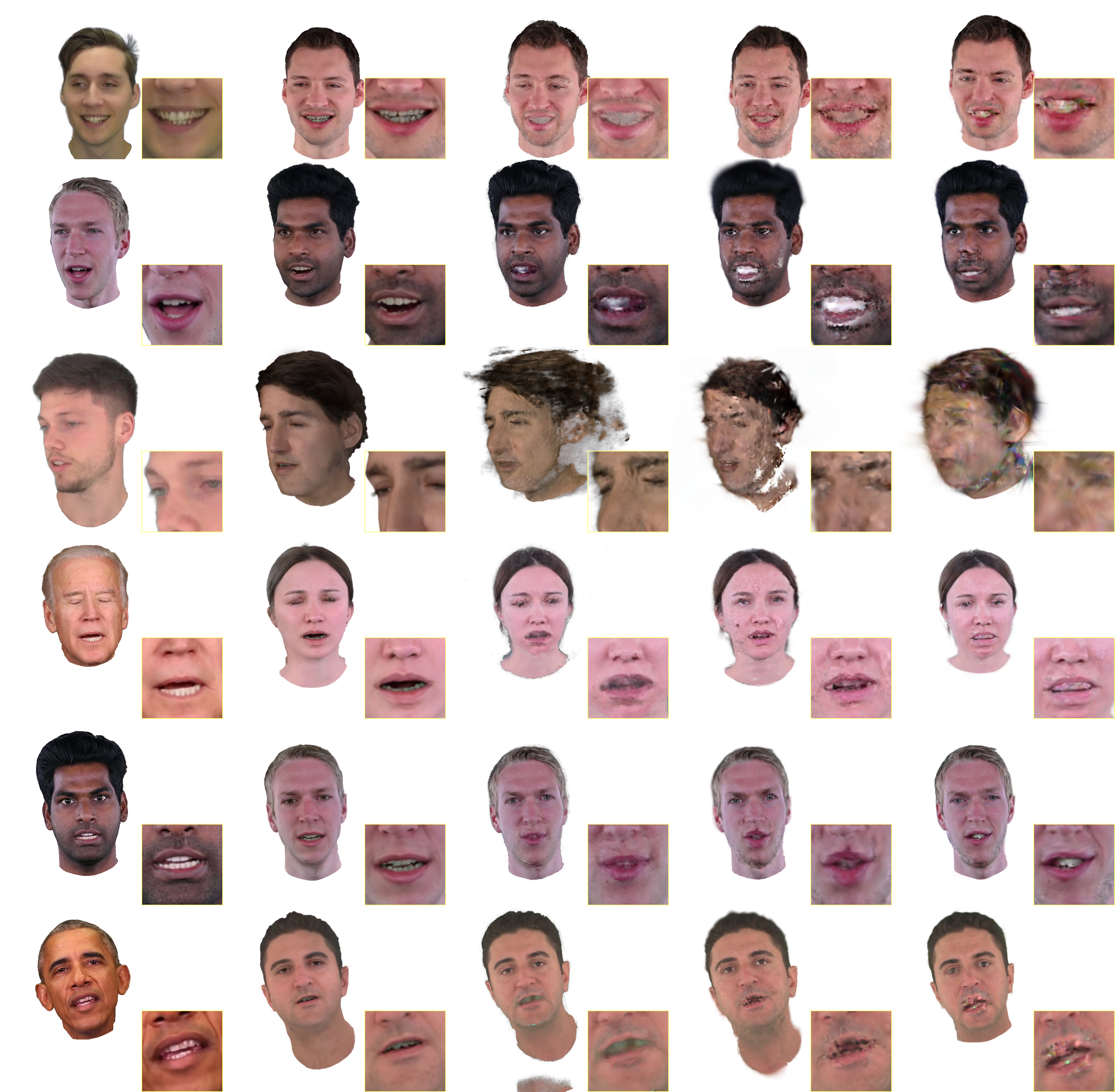}}
    \put(1.0, -0.4){Source}
    \put(3.0, -0.4){Ours}
    \put(4.8, -0.4){INSTA~\cite{Zielonka2022InstantVH}}
    \put(6.8, -0.4){SA \cite{shao2024splattingavatar}}
    \put(8.8, -0.4){FA \cite{xiang2024flashavatar}}
    \end{picture}
    \vspace{0.25cm}
    \caption{Cross-Reenactment on a Limited Number of Frames: We compare \shortname inversion using only 3 views to SOTA methods that utilize 13 frames. While the baseline methods produce good qualitative results on the test sequence with 13 frames, they all fail severely in novel view and expression evaluation.}
    \label{fig:transfer}
\end{figure*}

\end{document}